\PassOptionsToPackage{numbers,sort&compress}{natbib}
\documentclass{article}

\usepackage[eandd,final]{neurips_2026}
\usepackage[utf8]{inputenc} 
\usepackage[T1]{fontenc}    
\usepackage{hyperref}       
\usepackage{url}            
\usepackage{booktabs}       
\usepackage{amsfonts}       
\usepackage{nicefrac}       
\usepackage{microtype}      
\usepackage{xcolor}         
\usepackage{amsmath}
\usepackage{multirow}
\usepackage{graphicx}
\usepackage{subcaption}
\usepackage{algorithm}
\usepackage{algpseudocode}

\usepackage[disable]{todonotes}
\newcommand{\note}[1]{\todo[inline]{#1}}

\renewcommand{\note}[1]{}

\hypersetup{
colorlinks=true,
linkcolor={red!50!black},
filecolor=magenta,      
urlcolor={blue!80!black},
citecolor={blue!50!black},
pdftitle={ScAn-Bench: Evaluating Scaling Analysis Methodology}
}

\title{ScAn-Bench: Evaluating Scaling \\ Analysis Methodology}

\author{%
  \textbf{Artin Sermaxhaj}$^{1*}$ \quad
  \textbf{Nastaran Alipour}$^{1,2*}$ \quad
  \textbf{Donat Sinani}$^{1*}$ \quad
  \textbf{Johannes Hog}$^{1}$ \\
  \textbf{Neeratyoy Mallik}$^{1,2}$ \quad
  \textbf{Steven Adriaensen}$^{1}$ \quad
  \textbf{Jenia Jitsev}$^{3}$ \quad
  \textbf{Danny Stoll}$^{1}$ \\[0.5em]
  $^{1}$University of Freiburg,
  $^{2}$Zuse School ELIZA, \\
  $^{3}$Juelich Supercomputing Center (JSC),
  Research Center Juelich (FZJ) \\
  \texttt{\{sermaxha, alipourn, sinanid, stolld\}@cs.uni-freiburg.de}
}

\begin{document}

\maketitle

\begingroup
\renewcommand{\thefootnote}{\fnsymbol{footnote}}
\footnotetext[1]{Equal contribution. Ordering among equally contributing authors determined by roulette.}
\endgroup

\begin{abstract}




Recent progress in machine learning is driven by large-scale foundation models, where scaling laws and finding optimal scaling prescriptions for architecture, data, and hyperparameters are key in advancing the state-of-the-art. Therefore, it is surprising that no systematic study evaluates the methodology to obtain scaling laws and prescriptions across different model types. To shed light on this crucial blind spot and facilitate future research, we introduce the surrogate benchmarks ScAn-Bench-LLM and ScAn-Bench-VLM based on 4524 and 8024 checkpoints of language and vision-language model pipelines. On our benchmarks, we perform the first systematic evaluation of both data acquisition and extrapolation methodology for scaling analysis across different data modalities.
\end{abstract}

\section{Introduction}



The rapid advancement of foundation models has reshaped approaches to machine learning, driving progress in natural language processing~\cite{vaswani-neurips17a, brown-neurips20a, openai-arxiv23a, guo-nature2025a}, computer vision~\cite{radford-arxiv21a, oquab-tmlr24a}, and other fields.
As training at target scales is prohibitively expensive, researchers increasingly rely on scaling analysis (ScAn)~\cite{kaplan-arxiv20a, hoffmann-neurips22a, cherti-cvpr23a, nezhurina-neurips2025a, li-arxiv25a} to model empirical relationships between architecture, data, hyperparameters, compute, and loss in order to extrapolate performance, predict optimal training configurations, and derive scaling prescriptions.
Determining which configurations to evaluate, how to allocate compute across scales, and how to extrapolate reliably to larger regimes remains a central open problem for ScAn.
This challenge is especially acute for less-researched or emerging foundation‑model families, where, unlike for large language models (LLMs), established rules of thumb are largely absent. 


While recent studies have attempted to compare different ScAn methodologies, these evaluations are overwhelmingly focused on LLMs. More critically, these comparisons often suffer from empirical biases due to inconsistent meta-choices during model training, leading to discrepancies in the reported results even when using the same underlying methodology ~\cite{pearce2024reconcilingkaplanchinchillascaling, porian2025resolvingdiscrepanciescomputeoptimalscaling, li2025misfittingsurveyscalinglaws}. Existing evaluations also largely study extrapolation on static scaling datasets, while evaluating data acquisition strategies requires interactive and reproducible benchmarks.

In other meta-algorithmic fields such as neural architecture search (NAS) and hyperparameter optimization (HPO), which face similar challenges~\cite{yang-iclr20a}, tabular and surrogate benchmarks have emerged as key enabling research artefacts for efficient evaluation. By collecting a representative dataset of trained configurations, one can either precompute outcomes for a small search space exhaustively or fit a fast surrogate predictor over these data to approximate the outcome of expensive training runs in rich search spaces in seconds.
A prominent example of the former is NAS-Bench-101~\cite{ying-icml19a}, while surrogate benchmarks have since emerged as a widely adopted approach for large or continuous domains~\cite{siems-arxiv20a,eggensperger-neuripsdbt21a, yan-neurips21a, hirose-acml21a,zela-iclr22a,bansal-neurips22a,sukthanker-neurips24a}.

Thus, to enable the systematic study of ScAn methodology, we introduce the two surrogate benchmarks ScAn-Bench-VLM and ScAn-Bench-LLM, covering vision-language models (VLMs) and LLMs respectively.
We evaluate 9 scaling and hyperparameter settings over 1,000 training runs for each benchmark, resulting in 8,024 and 4,524 checkpoints for VLMs and LLMs, respectively, with model sizes ranging from 2M to 357M parameters for VLMs and from 16M to 1B for LLMs.
We fit this data to our surrogate predictor, modeling the training dynamics and downstream performance, enabling accurate approximation of training runs in CPU-seconds rather than GPU-hours.

We demonstrate that our benchmarks make it possible to study ScAn methods under reproducible conditions, presenting the first apples-to-apples comparison between existing methods
across multiple data modalities and foundation model families. Beyond our evaluation, our benchmarks will allow even researchers without access to large-scale GPU clusters to advance ScAn methodology.




Our main contributions are:
\begin{itemize}
    \item \textbf{Scaling analysis framework:} We develop a systematic view on scaling analysis to guide benchmarking and evaluation (Section \ref{sec:scaling_analysis}).
    
    \item \textbf{Surrogate benchmarks:}
    We introduce the first surrogate benchmarks (ScAn-Bench-LLM and ScAn-Bench-VLM) suitable for research on scaling analysis methodology (Section \ref{sec:surrogate_benchmarks}). 
    
    \item \textbf{Evaluation of scaling analysis methodology:} We provide the first controlled evaluation of both sequential data acquisition and extrapolation strategies across multiple foundation model families (Section \ref{sec:scan_methodology_evaluation}).
    
\end{itemize}

\note{DS: do we derive suggestions for practitioners as we promise in the abstract?}


Our code is available \href{https://github.com/automl/scan_bench/tree/neurips2026}{here}.


\section{A systematic view on scaling analysis}
\label{sec:scaling_analysis}
Existing scaling analysis methodologies differ not only in their extrapolation models, but also in how training configurations are sampled, compute is allocated, and optimization hyperparameters are chosen, and even what quantities are extrapolated~\cite{kaplan2020scaling,hoffmann-neurips22a,fetterman2023tunescalehyperparameteroptimization}. Recent studies have shown that these implementation choices can substantially affect empirical scaling conclusions~\cite{pearce2024reconcilingkaplanchinchillascaling, porian2025resolvingdiscrepanciescomputeoptimalscaling}. To enable systematic evaluation, we introduce a unified framework that decomposes scaling analysis into a \emph{data acquisition} phase under a constrained search budget and an \emph{extrapolation} phase that predicts optimal behavior at unseen compute scales.

\paragraph{The scaling problem}
We consider a pretraining task loss, $f$, to be minimized at a target compute budget $C_{\text{target}}$ (typically FLOPs). Given a joint hyperparameter search space $\Lambda = \Lambda_c \times \Lambda_o$, where $\Lambda_c$ is the space of compute-scaling hyperparameters and $\Lambda_o$ is the space of non-scaling optimization hyperparameters, our theoretical goal is to find the optimal configuration $(\lambda_c^*, \lambda_o^*)$ that minimizes the objective function subject to this final compute constraint. Note that the compute budget is strictly a function of the compute-scaling hyperparameters, $C(\lambda_c)$. This problem can be formally stated as:

\begin{equation} \label{eq:theoretical_objective}
\begin{aligned}
(\lambda_c^*, \lambda_o^*) = \;& \underset{(\lambda_c, \lambda_o) \in \Lambda}{\text{arg min}} \quad f(\lambda_c, \lambda_o) \quad \text{subject to} \quad C(\lambda_c) \leq C_{\text{target}}
\end{aligned}
\end{equation}

Scaling analysis methodology approximates this theoretical solution through a constrained data acquisition phase followed by an extrapolation phase. 

\paragraph{Data acquisition}
A data acquisition strategy samples a set of configurations and evaluates them to yield a dataset $\mathcal{D} = \{ \big(\lambda_{c,i}, \lambda_{o,i}, f(\lambda_{c,i}, \lambda_{o,i})\big) \}_{i=1}^M$. The sampling process is bounded by a total allotted search budget $\sum_{i=1}^{M} C(\lambda_{c,i}) \le C_{\text{search}}$, which in practice is usually not enough to run adequate evaluations on the target compute $C_{\text{target}}$.
A fundamental challenge in standardizing the evaluations is decoupling the data acquisition strategy from domain-specific heuristics. For instance, the original Chinchilla \cite{pearce2024reconcilingkaplanchinchillascaling} methodologies implicitly rely on heavily tuned prior knowledge for non-scaling hyperparameters, $\lambda_o$ (e.g., learning rate schedules). To ensure a rigorous, prior-free comparison of the allocation strategies, we must standardize how these baselines are instantiated.

\paragraph{Loss extrapolation} 
Utilizing the empirically acquired dataset $\mathcal{D}$, the most fundamental layer of extrapolation applies a predictive technique for the optimal loss. Rather than attempting to model the entire multi-dimensional loss landscape, the goal is to extrapolate the objective function $f$ along the empirically observed Pareto optimal frontier, directly predicting the minimal achievable loss for a given target compute budget $C_{\text{target}}$.

\paragraph{Coarse parameter extrapolation}
Beyond predicting raw loss, many existing scaling studies reduce problem complexity by treating the overall model size $N$ as a single scalar proxy for scale. This intermediate layer assumes that once the optimal $N$ is determined, the underlying hyperparameters can be derived via established domain heuristics \cite{pearce2024reconcilingkaplanchinchillascaling, kaplan2020scaling}. Under this regime, the extrapolation specifically models the model size $N^*$ and consequently optimal token count $D^*$.

\paragraph{Full extrapolation}
The most fine-grained layer of extrapolation eliminates the reliance on heuristic priors by providing a direct predictive prescription for the complete configuration. In this underexplored setting, the fully-parameterized extrapolation function, $\mathcal{Q}_{\Lambda}$, directly models the optimal joint set of compute-scaling and non-scaling optimization hyperparameters $(\hat{\lambda}_c^*, \hat{\lambda}_o^*)$ as a function of the target compute budget $C_{\text{target}}$.

\section{ScAn-Bench-LLM and ScAn-Bench-VLM}
\label{sec:surrogate_benchmarks}

Training and evaluating foundation models is computationally expensive, making direct evaluation of ScAn methods infeasible for many research labs, hindering progress in this area. To address this, we construct surrogate benchmarks that approximate the mapping between training configurations and performance. While surrogate predictions are obtained quickly on CPU, training based evaluation requires hours on GPUs and is orders of magnitude slower (see Table~\ref{tab:runtime_summary}). Unlike static tabular benchmarks, surrogate benchmarks can be queried outside the collected configurations, allowing for realistic design spaces and continuous parameters.

Section~\ref{sec:scan_spaces_data_collection} describes the ScAn spaces and data collection pipelines for VLM and LLM settings, while Section~\ref{sec:creating_surrogate_benchmarks} presents the surrogate modeling procedure. We provide additional details in Appendix~\ref{app:foundation_model_pipeline_details} and Appendix~\ref{app:surrogate_modeling_details}.

\begin{table}[t]
\centering
\caption{Runtime comparison for 100 evaluations. Surrogate querying is measured in CPU hours on AMD EPYC 9655 CPU using 8 cores, while training cost is reported in GPU hours on A100 GPUs.}
\label{tab:runtime_summary}
\begin{tabular}{lccc}
\toprule
Model family & Surrogate query (CPU-s) & Training (GPU-h) \\
\midrule
VLM   & 29 & 419.58 \\
LLM   & 17 & 3519.42 \\
\bottomrule
\end{tabular}
\end{table}

\subsection{ScAn spaces and data collection}
\label{sec:scan_spaces_data_collection}

We define two training pipelines with corresponding ScAn spaces for the considered model families.
The VLM pipeline trains CLIP~\cite{radford-arxiv21a} models using an open-source implementation~\cite{cherti-cvpr23a} on a 60M-sample subset of LAION-400M~\cite{schuhmann2021laion400mopendatasetclipfiltered}. The LLM pipeline trains decoder-only transformers~\cite{ajroldi-plainlm2024a} on the SlimPajama~\cite{soboleva-cerebras2023a} dataset.

\paragraph{ScAn space}
For each model family, we define a bounded ScAn space spanning both scaling variables and optimization hyperparameters. Table~\ref{tab:benchmark_summary} summarizes the number of continuous and discrete parameters considered across model families, while Appendix~\ref{app:scan_space} provides the full parameter ranges and domains. For VLMs, scaling variables include the vision and text encoder width together with the number of training samples. For LLMs, we vary the number of layers, number of attention heads, embedding dimension, and training tokens. Across all model families, we additionally vary key optimization hyperparameters. For VLMs, these include the learning rate, weight decay, warmup fraction, and optimizer coefficients $(\beta_1, \beta_2, \epsilon)$. For LLMs, we vary the learning rate, weight decay, cooldown steps, and optimizer coefficients $(\beta_1, \beta_2)$.

\paragraph{Data collection}
To capture training dynamics and allow queries at intermediate training stages, we recorded a total of 8024 checkpoints for VLMs and 4524 checkpoints for LLMs. To limit storage overhead, we employed a controlled checkpointing scheme that avoids excessive checkpoint density while preserving information of the training trajectory. Details on the checkpointing strategy are provided in Appendix~\ref{app:multi_fidelity_modeling_checkpoints}.
For each run, we record upstream and downstream performance, evaluating on all checkpoints.
We provide a complete overview of the recorded metrics and the model pipelines in Appendix~\ref{app:model_training_and_evaluation}.

\begin{table}[t]
\centering
\caption{Summary of the two surrogate benchmarks. The table reports the number of hyperparameters (\#HPs), scale parameters (\#Scale), FLOP ranges, continuous and discrete dimensions, and the number of downstream tasks (\#Down.) modeled by each surrogate benchmark.}
\label{tab:benchmark_summary}
\begin{tabular}{lccccccc}
\toprule
Benchmark & \#HPs & \#Scale & FLOP Range & Cont. & Disc. & \#Down. \\
\midrule
ScAn-Bench-VLM    & 6 & 3 & $3.6\times10^{13}$ -- $7.8\times10^{18}$ & 7 & 2 & 39 \\
ScAn-Bench-LLM    & 5 & 4 & $3.2\times10^{16}$ -- $2.9\times10^{20}$  & 6 & 3 & - \\
\bottomrule
\end{tabular}
\end{table}

\subsection{Creating the surrogate benchmarks}
\label{sec:creating_surrogate_benchmarks}

We follow existing literature on surrogate benchmark construction~\cite{sukthanker-neurips24a} for surrogate model candidates and evaluation procedures.

\paragraph{Surrogate modeling} We consider AutoGluon~\cite{erickson-arxiv20a}, XGBoost~\cite{chen-kdd16a}, LightGBM~\cite{ke-neurips17a}, TabPFN, and a mixed ensemble consisting of Linear Regression, Ridge Regression~\cite{hoerl-technometrics70a}, Random Forests~\cite{breiman-mlj01a}, XGBoost and LightGBM models. In addition, we also include TabPFN~\cite{grinsztajn-arxiv2026a} as a recent modeling approach for tabular setting, which we find to perform best. More information on each surrogate is provided in the Appendix~\ref{app:surrogate_candidates}.
\paragraph{Model selection}
Model selection is based on each surrogate's ability to predict upstream performance. Table~\ref{tab:surrogate_metrics_spearman_rmse} reports surrogate performance on the held out test set in terms of Spearman rank correlation and RMSE across all model families. TabPFN consistently yields the strongest performance across benchmarks and is therefore selected as the default surrogate predictor. Additional information on data split is provided in Appendix~\ref{app:surrogate_data_splitting}. A more comprehensive evaluation across additional metrics is provided in Appendix~\ref{app:surrogate_evaluation_metrics}.

\paragraph{Final model and runtime}
We use TabPFN as the predictor for both model families, trained on all available data, and model each target metric independently, including upstream and downstream metrics. Table~\ref{tab:benchmark_summary} summarizes the properties of the surrogate benchmarks. 
In addition to the performance surrogate, we train a separate binary surrogate for VLMs that predicts whether a configuration will diverge. Further details are provided in Appendix~\ref{app:divergence_prediction}. The predictors are integrated into a broader API that exposes additional configuration-level information (Table~\ref{tab:api_interface}), with further details in Appendix~\ref{app:surrogate_api}.

\begin{table}[t]
\centering
\small
\caption{Comparison of surrogate performance across model families in predicting upstream performance, measured by Spearman rank correlation (higher is better) and RMSE (lower is better). }
\label{tab:surrogate_metrics_spearman_rmse}
\begin{tabular}{lcc cc cc}
\toprule
& \multicolumn{2}{c}{VLM} 
& \multicolumn{2}{c}{LLM} \\

\cmidrule(lr){2-3}
\cmidrule(lr){4-5}
\cmidrule(lr){6-7}

Surrogate
& Spearman $\uparrow$ & RMSE $\downarrow$
& Spearman $\uparrow$ & RMSE $\downarrow$ \\
\midrule

TabPFN         
& \textbf{0.98} & \textbf{0.21}
& \textbf{0.95} & \textbf{0.27} \\

AutoGluon      
& 0.98 & 0.26
& 0.91 & 0.30 \\

Ensemble (XGB) 
& 0.96 & 0.35
& 0.85 & 0.35 \\

Ensemble (Mix) 
& 0.95 & 0.37
& 0.86 & 0.33 \\

Ensemble (LGB) 
& 0.96 & 0.42
& 0.87 & 0.33 \\

\bottomrule
\end{tabular}
\end{table}


\section{A systematic evaluation of ScAn methodology}
\label{sec:scan_methodology_evaluation}

As introduced in Section \ref{sec:scaling_analysis}, scaling analysis requires two distinct methodological choices: a data acquisition strategy to gather data at lower compute scales, and an extrapolation strategy to project those observations to a target compute regime. To systematically benchmark these components, we evaluate several standard approaches from the literature for both phases. Using the ScAn-Bench surrogate benchmarks introduced in Section~\ref{sec:surrogate_benchmarks}, we conduct a scaling analysis to answer the following questions:

\begin{itemize}
    \item[$RQ1$:] What are the trade-offs during data acquisition among maximizing immediate performance, broadly exploring the Pareto frontier, and ensuring a highly accurate scaling law fit?
    
    \item[$RQ2$:] Is there a combination of loss extrapolation technique and data acquisition that consistently performs well across benchmarks?
    
    \item[$RQ3$:] Is there a combination of coarse parameter extrapolation technique and data acquisition that consistently performs well across benchmarks?
    
    \item[$RQ4$:] Does the studied full extrapolation approach provide reliable and accurate results?

    \item[$RQ5$:] Do different acquisition approaches systematically favor particular extrapolation techniques?
\end{itemize}

\subsection{Evaluated ScAn Methodology}

\paragraph{Data acquisition strategies}
As established in Section \ref{sec:scaling_analysis}, decoupling strategies from domain-specific heuristic is essential to have a valid comparison between strategies across different domains. To this end, we adapt four generalized heuristic-free approaches. From the Chinchilla \cite{hoffmann-neurips22a} study, we evaluate Iso-Parameter Profiling (Chinchilla Approach 1) and Iso-FLOP Profiling (Chinchilla Approach 2), both of which rely on deterministic, grid-based sampling strategies. To evaluate dynamic sampling, we include Cost-Aware Robust Bayesian Search (CARBS) ~\cite{fetterman2023tunescalehyperparameteroptimization}, which utilizes an adapted EI acquisition function to sequentially sample configurations based on previous observations. Finally, we include Random Search as a baseline to isolate the performance gains of the structured allocation strategies. The design choices for these approaches can be found in Appendix \ref{app:scan_method_design_choices}.


\paragraph{Loss extrapolation} The coarsest level, which directly predicts the minimal loss at the target compute scale using either a Kaplan-style power law fit \cite{pearce2024reconcilingkaplanchinchillascaling} or a Chinchilla-style parametric loss function \cite{hoffmann-neurips22a}.

\paragraph{Coarse parameter extrapolation} This level predicts the optimal compute-scaling parameters, specifically the model size ($N^*$) and the token count ($D^*$). Here, we evaluate deriving $N^*$ and $D^*$ via the Chinchilla parametric loss estimate under cost constraints, as well as via direct power law fitting between the macro-parameters and total compute.

\paragraph{Full extrapolation} The most fine-grained level that models both $\lambda_c^*$ and $\lambda_o^*$ at the target scale. As this setting remains largely unexplored, we include a representative baseline based on independent linear regression (as in CARBS), leaving more expressive joint approaches for future work.


\subsection{Experimental setup}

\paragraph{Hyperparameter space}
Our data acquisition phase is strictly confined to a lower-compute regime bounded by $C_{\text{max}}$. To rigorously evaluate the performance of the extrapolation techniques, we define the unobserved target compute budget as $C_{\text{target}} = 20 \times C_{\text{max}}$. By anchoring $C_{\text{target}}$ within the compute scale of our benchmark datasets, we ensure the availability of ground-truth evaluations at the target scale. This prevents reliance on an unverified extrapolative regime, allowing us to precisely measure the true extrapolation error. While our overall hyperparameter search space $\Lambda$ is defined in compliance with these existing benchmarks, the specific configurations sampled during data acquisition are strictly restricted to configurations costing less than $C_{\text{max}}$. Specifically, we establish $C_{\text{max}} = {2 \times 10^{16}}$ FLOPs for the OpenCLIP benchmark and $C_{\text{max}} = {1.0 \times 10^{19}}$ FLOPs for the LLM benchmark.

\paragraph{Loss function} In line with scaling law literature \cite{kaplan2020scaling, hoffmann-neurips22a}, we establish pretraining validation loss as our objective function, $f$. While downstream task performance is frequently used for general model evaluation, standard extrapolation methodologies are predominantly formulated around validation loss. This is because validation loss provides a monotonically decreasing signal that is well-suited to power-law fitting and parametric modeling. By strictly isolating validation loss, we ensure our methodological evaluation is decoupled from the unpredictable dynamics of downstream tasks.
\paragraph{Seed Aggregation} To mitigate the impact of variance from our random hyperparameter initialization, we conducted the experiment in 10 independent seeds. Throughout Section \ref{sec:results}, we report the mean performance across these runs, along with the standard error to quantify predictive stability.

\subsection{Results}
\label{sec:results}
\begin{figure*}[htbp]
    \centering
    \includegraphics[width=\textwidth]{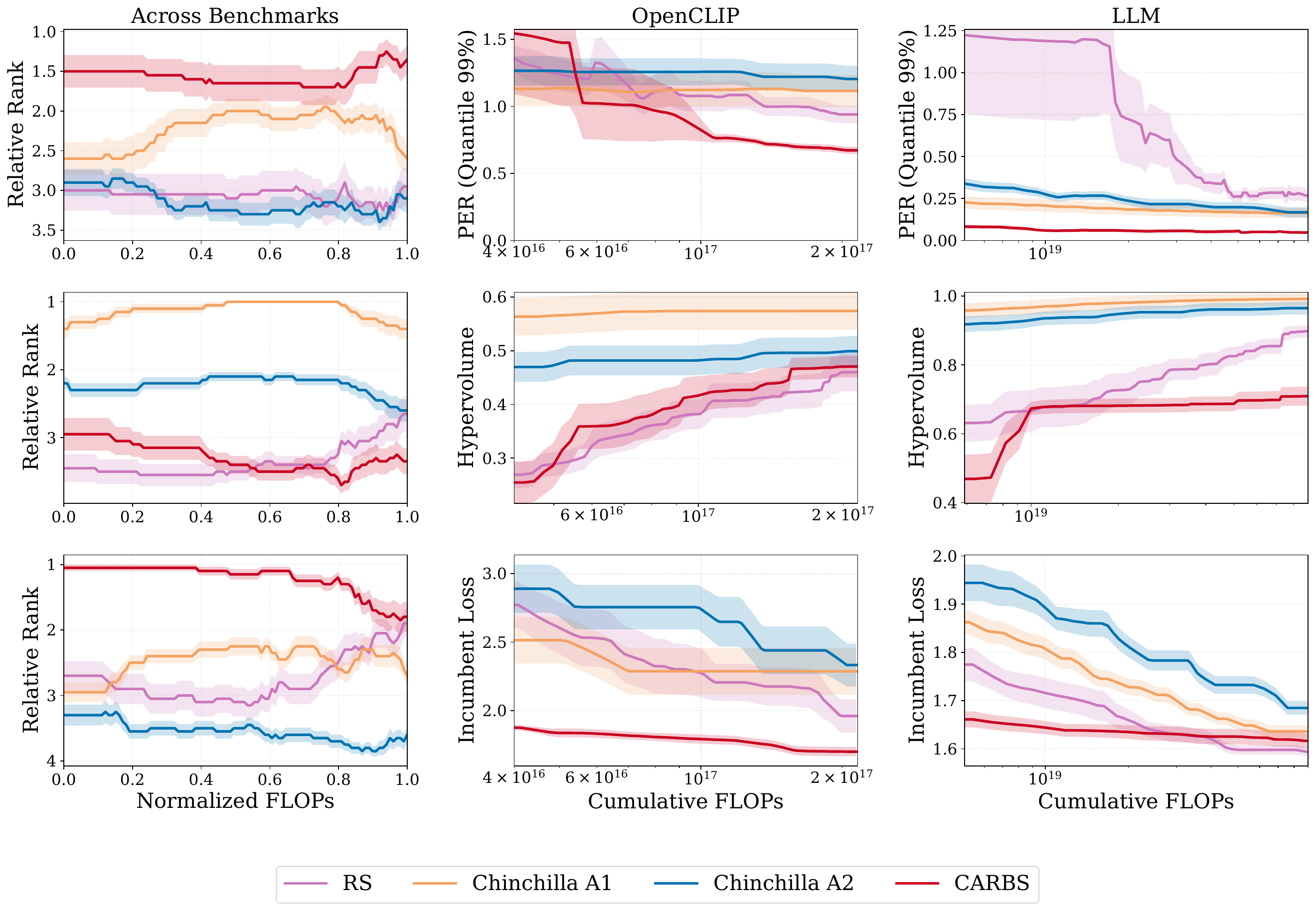}
    \caption{\textbf{Standalone evaluation of data acquisition strategies.} The performance of 4 different data acquisition strategies within the search space across three metrics against the cumulative search budget: PER (top, measuring proximity to the true optimal frontier), Hypervolume (middle, measuring broad spatial exploration), and Incumbent Loss (bottom, tracking the best loss found).}
    \label{fig:acquisition_metrics}
\end{figure*}

\begin{figure*}[htbp]
    \centering
    \includegraphics[width=0.75\textwidth]{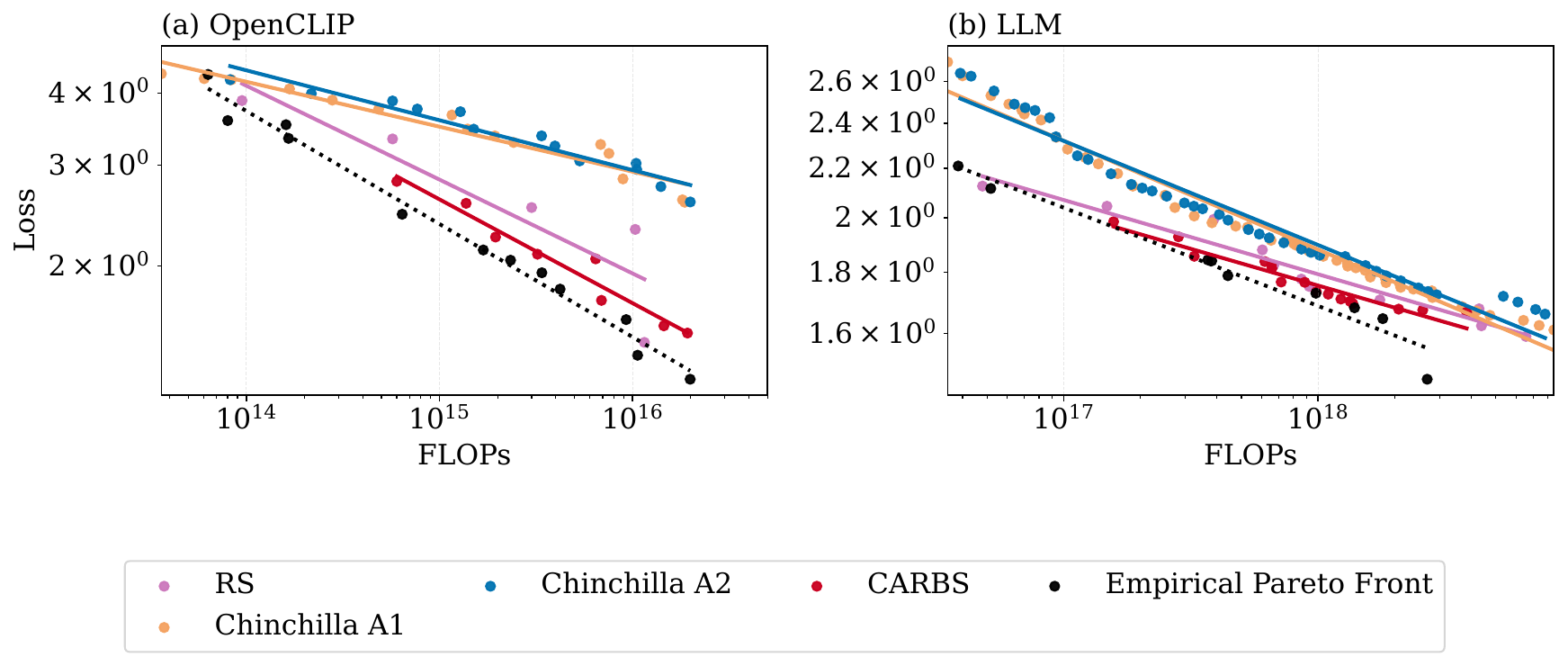}
    \caption{\textbf{Power law fit across acquisition strategies.} Projected Power-Law fits for (a) OpenCLIP and (b) LLM for a single selected random seed, evaluated at the maximum search budget and plotted alongside the true empirical Pareto optimal frontier.
    }
    \label{fig:single_seed_pareto_front}
\end{figure*}

\paragraph{Data acquisition}
As established at the start of Section~\ref{sec:scan_methodology_evaluation} the primary objective of this evaluation is to systematically analyze the trade-offs between different data acquisition strategies and prediction methodologies. In doing so, we demonstrate how notoriously sensitive scaling law projections are to underlying design choices.
Figure~\ref{fig:acquisition_metrics} evaluates performance across three metrics: Pareto Estimation Regret (PER, top) measures the area between the acquired data's fit and the true Pareto frontier, Hypervolume (middle) measures broad spatial coverage and Incumbent Loss (bottom) tracks the best loss achieved over the search trajectory. 
Observing the results, a clear exploration-exploitation trade-off emerges. CARBS efficiently exploits high-performing regions, achieving the lowest PER and superior early-stage Incumbent Loss. However, the deterministic Iso-Parameter approach ensures broad exploration, yielding the highest Hypervolume. Tracking the best loss found as the compute budget expands, RS surpasses CARBS across benchmarks. This suggests that while CARBS excels early, its exploitation bias restricts its search space during later acquisition phases. Figure \ref{fig:single_seed_pareto_front} plots the scaling law fits and the empirical Pareto frontier for a single seed, directly illustrating the localized nature of the data acquired via CARBS.

\note{plots with new labels for chinchilla A1, A2 and Incumbent Loss need to be added}

\paragraph{Loss extrapolation}
We assess loss-level extrapolation techniques under both Iso-Parameter Profiling (Chinchilla A1) and CARBS data acquisition by examining their predicted loss at $C_{\text{target}}$ as a function of cumulative compute. Our findings indicate that no single data acquisition and extrapolation combination is optimal across all benchmarks. 

\begin{figure*}[htbp]
    \centering
    \includegraphics[width=0.75\textwidth]{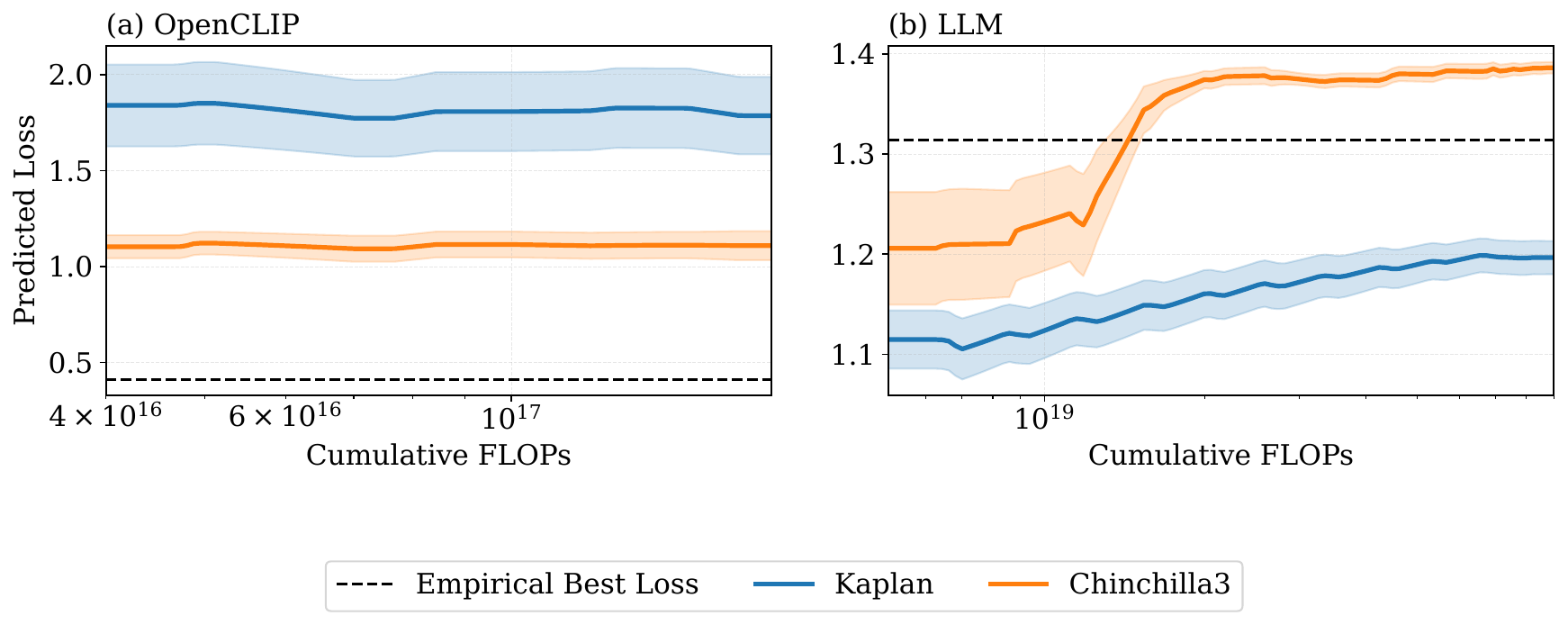}
    \caption{\textbf{Comparison of loss extrapolation techniques under Iso-Parameter Profiling (Chinchilla A1).} Predicted loss trajectories over cumulative FLOPs for the \textbf{(a)} OpenCLIP and \textbf{(b)} LLM benchmarks. The plots compare the Chinchilla Approach 3 and Kaplan prediction methods against the empirical true optimal loss (dashed line) at $C_{\text{target}}$. While the Chinchilla technique generally achieves better overall performance across the trajectory, the Kaplan method demonstrates a more stable, monotonic improvement in its predictions as the compute budget scales over time.}
    \label{fig:loss_predictor_c1_comparison}
    \vspace{2em}
    \centering
    \includegraphics[width=0.75\textwidth]{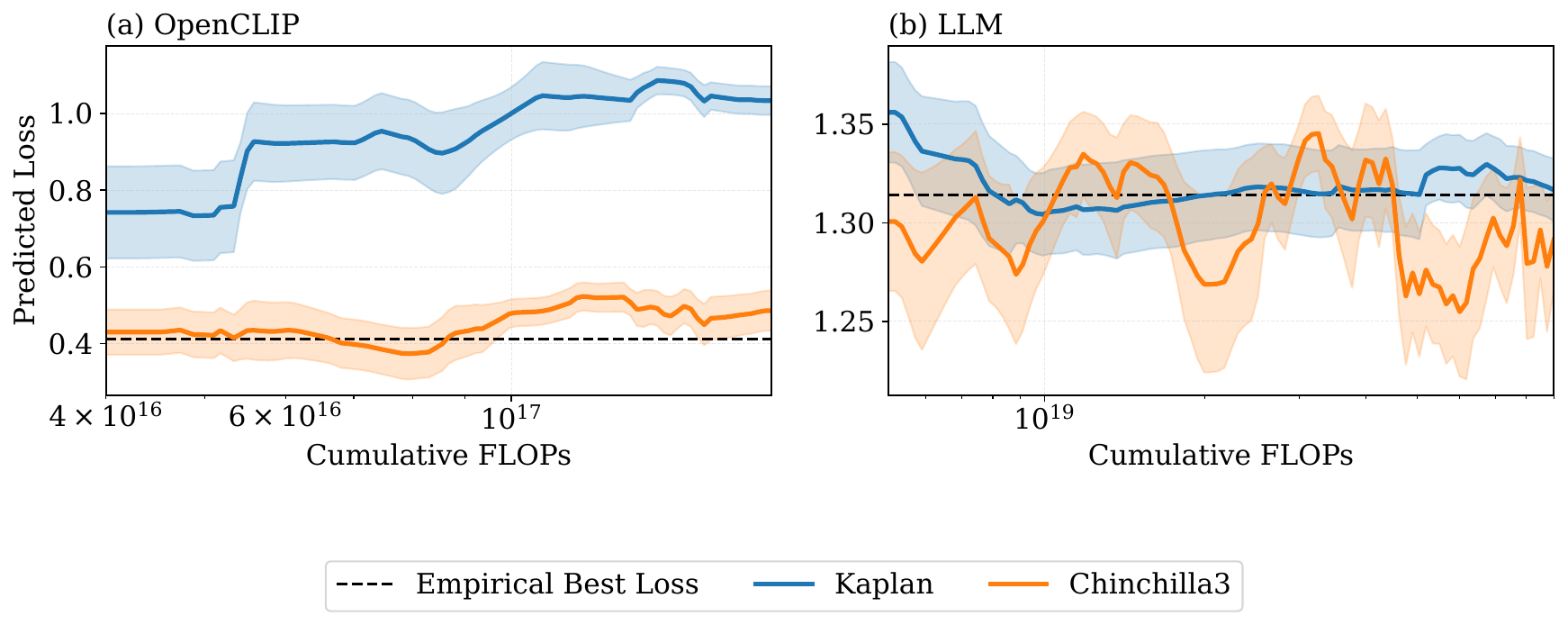}
    \caption{\textbf{Comparison of loss extrapolation techniques under CARBS data acquisition.} Predicted loss trajectories over cumulative FLOPs against the empirical true optimal loss (dashed line) for the \textbf{(a)} OpenCLIP and \textbf{(b)} LLM benchmarks.}
    \label{fig:carbs_error}
\end{figure*}

Under Iso-Parameter Profiling (Figure~\ref{fig:loss_predictor_c1_comparison}, Chinchilla A1), the parametric loss model (Chinchilla Approach 3) achieves closer predictions to the empirical target loss, while Kaplan exhibits more stable, monotonic improvement with increasing compute. Figure~\ref{fig:carbs_error} illustrates this same comparison when data is acquired via the CARBS strategy. Chinchilla Approach 3 significantly outperforms Kaplan in anytime prediction on the OpenCLIP benchmark. Conversely, Kaplan yields more reliable and accurate predictions on the LLM benchmark under this acquisition regime.

\paragraph{Coarse parameter extrapolation}

\begin{figure*}[htbp]
    \centering
    \includegraphics[width=0.75\textwidth]{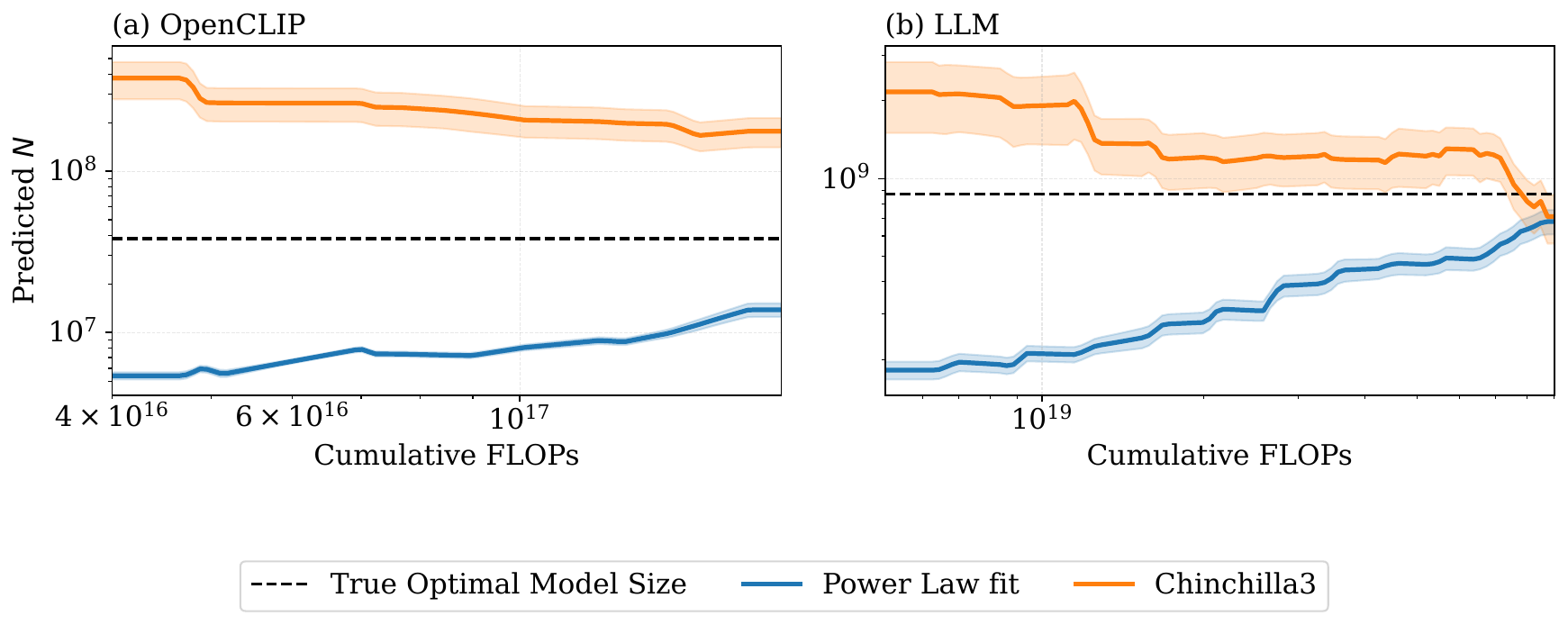}
    \caption{\textbf{Model size extrapolation under Iso-Parameter data acquisition.} Trajectories of the predicted optimal model size ($N^*$) for the \textbf{(a)} OpenCLIP and \textbf{(b)} LLM benchmarks. The parametric Chinchilla approach fails on OpenCLIP due to invalid compute estimation heuristics, whereas the purely empirical Power-Law fit remains robust across both domains.}
    \label{fig:comparision_c1}
    
    \vspace{2em} 
    
    \includegraphics[width=0.75\textwidth]{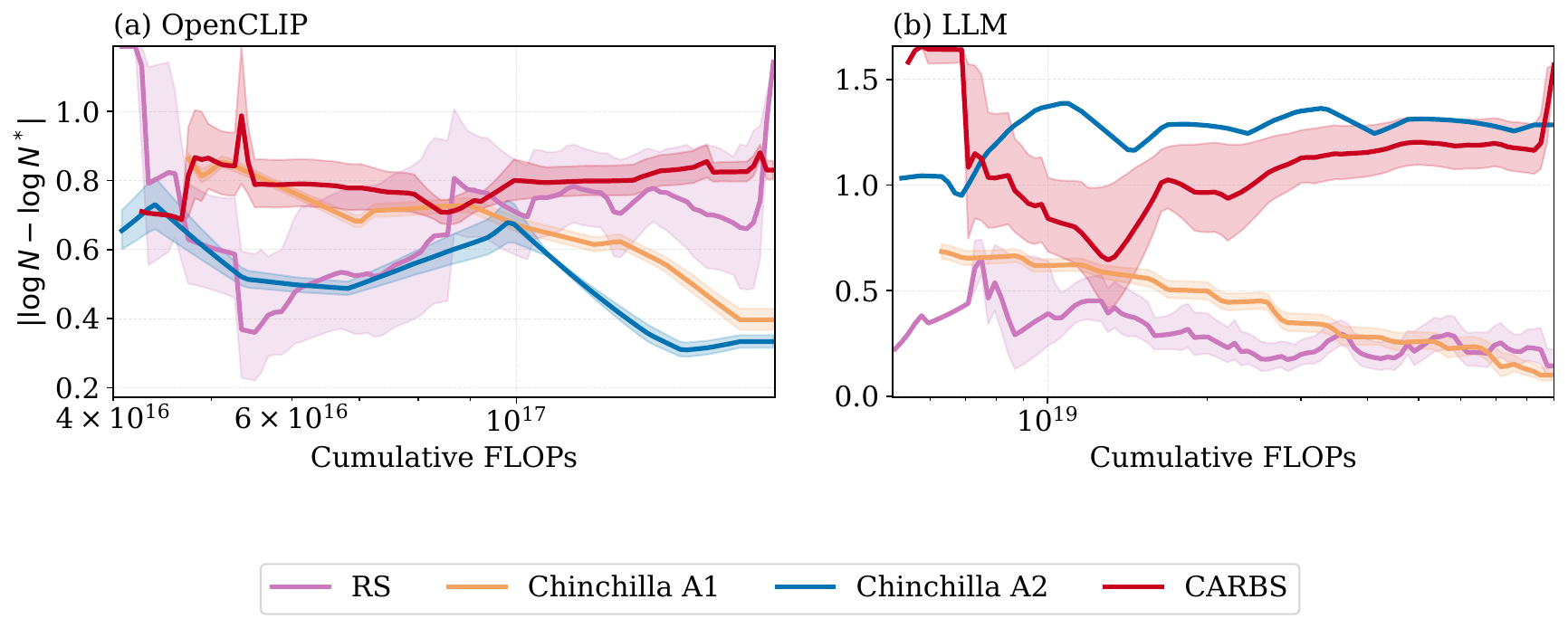}
    \caption{\textbf{Robustness of Power-Law model size predictions across acquisition strategies.} The y-axis represents the absolute prediction error, calculated as the absolute difference between the log-scale of the predicted model size and the true empirical optimal model size ($| \log_{10}(N_{\text{pred}}) - \log_{10}(N_{\text{true}}^*) |$) at the target compute budget $C_{\text{target}}$. With the exception of CARBS, most of the profiling strategies facilitate accurate predictions of $N^*$ at the given search budget.}
    \label{fig:power_law_n_param_composite}
\end{figure*}

Parametric extrapolation models are fundamentally brittle when their underlying compute assumptions are violated. A primary limitation of using the Chinchilla Approach 3 to model optimal model size ($N^*$) is its strict reliance on the standard $C \approx 6ND$ compute estimation. To strictly evaluate the original formulation without injecting novel correction heuristics for multi-modal architectures, we held this standard compute estimation constant across all benchmarks. As demonstrated in Figure~\ref{fig:comparision_c1}(a) under Iso-Parameter profiling, this rigid assumption causes the Chinchilla technique to fail in predicting the optimal OpenCLIP model size. Conversely, Figure~\ref{fig:comparision_c1}(b) confirms that both the Chinchilla and empirical Power-Law projection methods reliably predict $N^*$ for the standard LLM benchmark. This contrast highlights that the purely empirical Power-Law Fit is a  more robust and generalized estimator across diverse architectural domains. We therefore isolate the performance of the Power-Law approach across data acquisition strategies in Figure~\ref{fig:power_law_n_param_composite}. Except for CARBS, which yields poor predictions due to its localized exploitation of the search space, other acquisition methodologies have accurate predictions of $N^*$ at the given search budget.

\paragraph{Full parameter extrapolation}
Predicting the full set of hyperparameters for a given compute budget, $C_{\text{target}}$, represents the most direct solution to the scaling law problem. As detailed in Section \ref{sec:scan_methodology_evaluation}, the specific full parameter extrapolation strategy we evaluate involves fitting independent linear regression models to each hyperparameter dimension. This regression fits the empirically observed Pareto optimal points as inputs, a methodology used by CARBS \cite{fetterman2023tunescalehyperparameteroptimization}. Table \ref{tab:actual_loss_comparision} compares the actual loss achieved by these predicted configurations $(\hat{\lambda}_c^*, \hat{\lambda}_o^*)$ against the best empirical loss at $C_{\text{target}}$. 

For this extrapolation experiment, we isolate our analysis to CARBS and Random Search acquisition strategies as they dynamically sample the non-scaling hyperparameters ($\lambda_o$) during the acquisition phase, whereas the deterministic grid-based methods hold $\lambda_o$ fixed across a given experiment. We evaluate the prediction quality across varying search budgets, specifically $C_{\text{search}} \in \{0.1, 0.5\} \times C_{\text{target}}$. As shown in Table \ref{tab:actual_loss_comparision}, as the search budget increases, the standard error across different random seeds decreases, indicating more stabilized extrapolation over time. Notably, Random Search outperforms CARBS across all evaluated budgets. Furthermore, the accuracy of the Random Search predictions monotonically improves by yielding lower actual losses as more data is acquired. Crucially, this approach is highly sensitive to the exact composition of the empirical Pareto frontier. This naive method assumes dimensional independence, ignoring the coupled scaling correlations between hyperparameters. For example, the predicted OpenCLIP configuration failed to account for interactions between the text and image tower hyperparameters, resulting in severely degraded performance.


\begin{table}[t]
\centering
\setlength{\tabcolsep}{4pt}
\caption{\textbf{Actual loss at predicted configurations.} Evaluation of full-parameter extrapolations using CARBS and Random Search (RS) at various $C_{\text{target}}$ budget ratios.}
\label{tab:actual_loss_comparision}
\begin{tabular}{lcccccc}
\toprule
& \multicolumn{3}{c}{LLM} & \multicolumn{3}{c}{OpenCLIP} \\
\cmidrule(lr){2-4} \cmidrule(lr){5-7}
Strategy
& 0.1x & 0.5x & Best
& 0.1x & 0.5x & Best \\
\midrule
CARBS
& $2.02 \pm 0.23$
& $\mathbf{1.76 \pm 0.17}$
& $1.31$
& $2.92 \pm 0.41$
& $2.49 \pm 0.14$
& $0.41$ \\
RS
& $\mathbf{1.71 \pm 0.18}$
& $1.89 \pm 0.37$
& $1.31$
& $\mathbf{2.28 \pm 0.29}$
& $\mathbf{1.81 \pm 0.30}$
& $0.41$ \\
\bottomrule
\end{tabular}
\end{table}

\begin{figure*}[htbp]
    \centering
    \includegraphics[width=\textwidth]{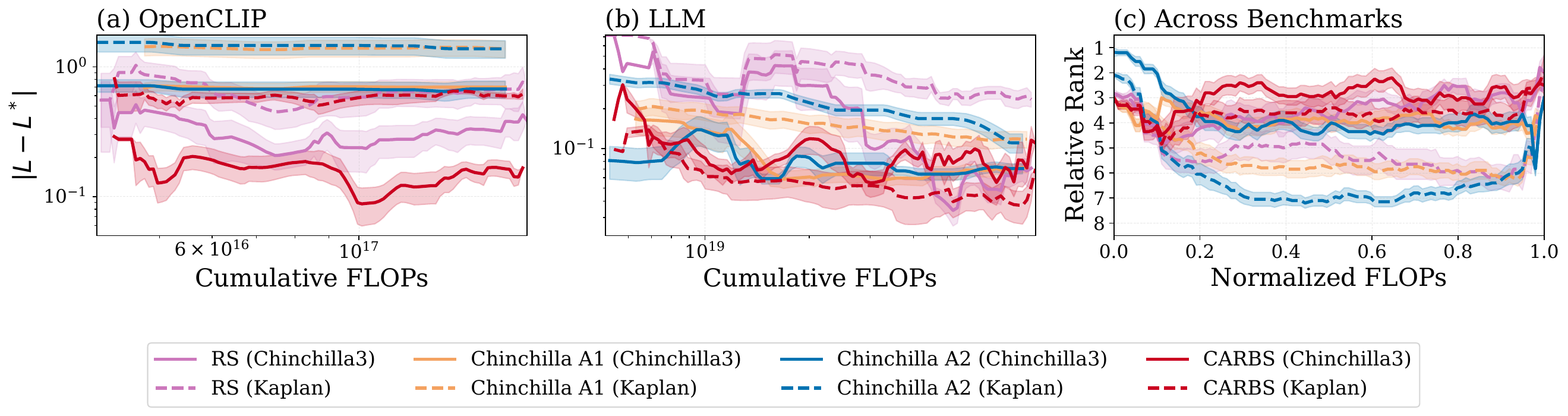}
    \caption{\textbf{Interaction between data acquisition strategies and loss prediction methods.} This figure illustrates the interplay between acquisition strategies and prediction techniques (Chinchilla 3 and Kaplan) across the \textbf{(a)} OpenCLIP and \textbf{(b)} LLM benchmarks, alongside \textbf{(c)} their relative performance rankings.}
    \label{fig:20x_loss_composite}
    \vspace{1em}
    \centering
    \includegraphics[width=\textwidth]{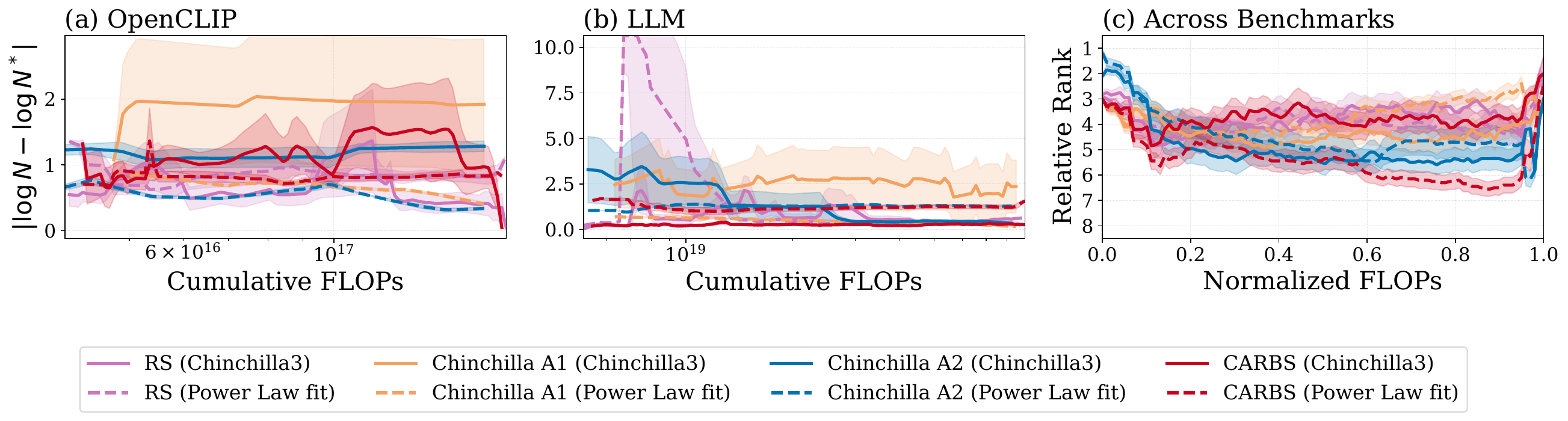}
    \caption{\textbf{Interaction between data acquisition strategies and model size prediction methods.} This figure illustrates the interplay between acquisition strategies and prediction techniques (Chinchilla 3 and Power Law fit) across the \textbf{(a)} OpenCLIP and \textbf{(b)} LLM benchmarks, alongside \textbf{(c)} their relative performance rankings.}
    \label{fig:20x_n_param_composite}
\end{figure*}

\paragraph{Coupled dynamics of data acquisition and extrapolation}
To answer research question $RQ5$ directly, we conducted an experiment on a grid of all acquisition and extrapolation strategies. With respect to the loss extrapolation, the parametric model (Chinchilla 3) consistently outperforms Power-Law fit (Kaplan) (Figure \ref{fig:20x_loss_composite}). This advantage stems from its flexible five-parameter formulation, which effectively captures the irreducible loss and accurately fits the acquired data. However, this flexibility introduces identifiability challenges: the model can fit the loss surface without isolating the true data-generating parameters. Consequently, this success does not translate to optimal model size ($N^*$) prediction. Because the parametric derivation of $N^*$ strictly depends on the precise ratio of specific fitted coefficients, it is highly brittle under poor identifiability. Instead, the efficacy of $N^*$ prediction is dictated by whether the data acquisition phase explicitly traces a usable compute-optimal frontier. Structured profiling methods (Chinchilla A1/A2) are designed to map this envelope, making the Power-Law fit highly accurate and efficient. Conversely, unstructured methods like Random Search broadly scatter $(N, D)$ allocations without targeting the envelope, while highly exploitative methods like CARBS sample too narrowly. These traces fail to map the Pareto frontier, making the Power-Law approach less effective and leaving the parametric fit, despite its identifiability caveats, as the only remaining viable option (Figure \ref{fig:20x_n_param_composite}; see Appendix~\ref{app:ablation} for comprehensive surrogate and corpus-split ablations).

\section{Related benchmarks and evaluations}
\label{sec:related_benchmarks}

\paragraph{Benchmarks} A large body of work studies benchmarking for architecture search ~\cite{ying-icml19a, dong-iclr20a, siems-arxiv20a, zela-iclr20b, zela-iclr22a, yan-neurips21a, klyuchnikov-ieee22a, mehrotra-iclr21a, dong-tpami21a, li-iclr21a, duan-cvpr21a}, hyperparameter optimization ~\cite{eggensperger-bayesopt13a, eggensperger-aaai15a, eggensperger-neuripsdbt21a, turner2021bayesianoptimizationsuperiorrandom, bayesmark}, and joint hyperparameter and architecture search~\cite{klein-nas_hpo_bench21a, zimmer-tpami21a, hirose-acml21a, bansal-neurips22a}. These benchmarks typically focus on smaller MLP or CNN architectures, often assume fixed training budgets, or do not provide suitable scaling parameters. Recent work has explored transformer-based models, but does not model independent training runs or hyperparameter variation~\citep{sukthanker-neurips24a}. 
To support benchmarking ScAn methods, we provide rich scale and hyperparameter spaces for VLM and LLM settings, support querying upstream metrics, downstream task performance, and report other necessary statistics such as FLOPs and parameter count. 

\paragraph{Evaluations in the language domain}
For LLMs, recent efforts have begun to evaluate the methods used to derive scaling laws. For instance, \citet{porian2025resolvingdiscrepanciescomputeoptimalscaling}  reveal that discrepancies in optimal scale predictions often stem from inconsistent parameter and compute counting heuristics. Similarly, \citet{choshen2025hitchhikersguidescalinglaw} highlight the high variance and methodological pitfalls of extrapolating scaling laws from intermediate training checkpoints rather than fully converged models. As for data-constrained regimes, \citet{muennighoff2025scalingdataconstrainedlanguagemodels} isolate the data axis, demonstrating that training under these regimes with repeated corpora yields similar power-law dynamics, albeit with diminishing returns. Moving beyond scalar model size, \citet{mcleish2025gemstonesmodelsuitemultifaceted} investigate the impact of architectural shape over the compute horizon, however, their evaluation remains restricted to a highly constrained pool of model sizes.
\note{DS: Does the gemstone paper do any evaluation of scan methodology? We should relate to the existing work alongside that aspect.}

\section{Conclusion}\label{sec:conclusion}


To facilitate the systematic study of ScAn methodology and make it computationally feasible, we introduced the two surrogate benchmarks, ScAn-Bench-VLM and ScAn-Bench-LLM, and evaluated existing scaling methodologies. We hope ScAn-Bench enables the community to develop and evaluate new ScAn methodology in a more systematic and reproducible manner, and that the benchmarks we provide serve as a template and encourage others to create additional benchmarks.

\paragraph{Limitations} Our benchmarks provide an initial framework for studying the scaling problem,
the compute scales, however, remain substantially smaller than current state-of-the-art models. Consequently, the observed trends may not fully transfer to significantly larger-scale regimes. 
We do not evaluate the sensitivity of the ScAn methods to their hyperparameters. 
VLM and LLM pipelines have substantially different characteristics, so, where results align across these settings, this provides some evidence that the finding may extend beyond a single pipeline.  However, as our evaluation also shows, results do not always align, and they may not even align across different architectures and training recipes for the same data modality. Since  evaluations of ScAn methodology typically consider only a single setting, this highlights the risk of drawing overly general conclusions from one pipeline. Similar to how benchmarking progressed in other meta-algorithmic fields, we hope our work will motivate future benchmarks across additional model families, architectures, and training recipes.
Finally, while we restrict our analysis to upstream scaling behavior, our VLM benchmark includes downstream task metrics.


\section*{Acknowledgments}
This research was funded by the Deutsche Forschungsgemeinschaft (DFG, German Research Foundation) under grant number 539134284, through EFRE (FEIH\_2698644) and the state of Baden-Württemberg. The authors gratefully acknowledge the computing time made available to them on the high-performance computers and at the NHR Centers at TU Dresden and KIT. These centers are jointly supported by the Federal Ministry of Research, Technology and Space of Germany and the state governments participating in the NHR (\url{www.nhr-verein.de/unsere-partner}). The authors gratefully acknowledge the Gauss Center for Supercomputing eV (\url{www.gauss-centre.eu}) for funding this project by providing computing time through the John von Neumann Institute for Computing (NIC) on the GCS Supercomputer JUWELS at Jülich Supercomputing Center (JSC) . The authors acknowledge funding by the European Union (via ERC Consolidator Grant DeepLearning 2.0, grant no.~101045765). Views and opinions expressed are however those of the author(s) only and do not necessarily reflect those of the European Union or the European Research Council. Neither the European Union nor the granting authority can be held responsible for them. Neeratyoy Mallik and Nastaran Alipour are supported by the Konrad Zuse School of Excellence in Learning and Intelligent Systems (ELIZA) through the DAAD programme ``Konrad Zuse Schools of Excellence in Artificial Intelligence'', sponsored by the Federal Ministry of Education and Research. Johannes Hog acknowledges funding by the Deutsche Forschungsgemeinschaft (DFG, German Research Foundation) under SFB 1597 (SmallData), grant number 499552394.

\begin{center}
    \begin{minipage}{0.6\textwidth} %
        \centering
    \includegraphics[width=0.45\linewidth]{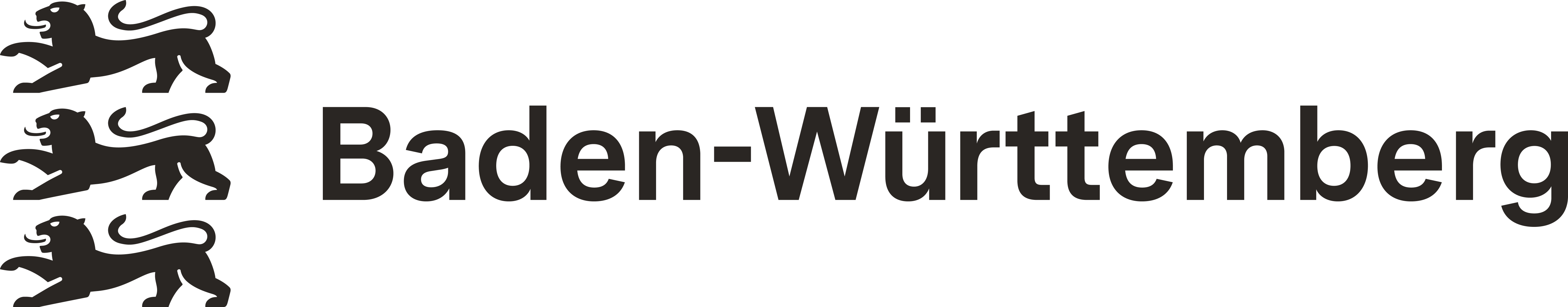}%
        \hfill
    \includegraphics[width=0.45\linewidth]{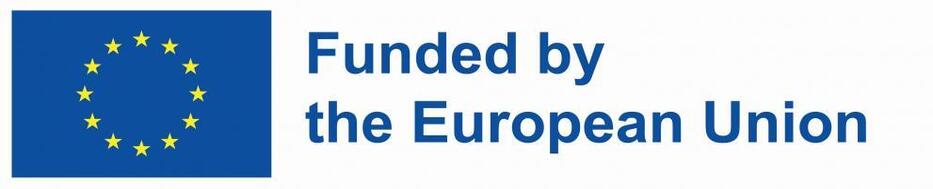}
    \end{minipage}
\end{center}

\bibliographystyle{unsrtnat} 
\bibliography{references, lib, strings, proc}

\newpage
\section*{NeurIPS Paper Checklist}

\begin{enumerate}

\item {\bf Claims}
    \item[] Question: Do the main claims made in the abstract and introduction accurately reflect the paper's contributions and scope?
    \item[] Answer: \answerYes{} 
    \item[] Justification: See Section~\ref{sec:scaling_analysis} for the scaling analysis methodology, Section~\ref{sec:surrogate_benchmarks} for surrogate benchmark construction, Section~\ref{sec:scan_methodology_evaluation} for the evaluation of ScAn methodologies, and Section~\ref{sec:results} for the experimental results.
    \item[] Guidelines:
    \begin{itemize}
        \item The answer \answerNA{} means that the abstract and introduction do not include the claims made in the paper.
        \item The abstract and/or introduction should clearly state the claims made, including the contributions made in the paper and important assumptions and limitations. A \answerNo{} or \answerNA{} answer to this question will not be perceived well by the reviewers. 
        \item The claims made should match theoretical and experimental results, and reflect how much the results can be expected to generalize to other settings. 
        \item It is fine to include aspirational goals as motivation as long as it is clear that these goals are not attained by the paper. 
    \end{itemize}

\item {\bf Limitations}
    \item[] Question: Does the paper discuss the limitations of the work performed by the authors?
    \item[] Answer: \answerYes{} 
    \item[] Justification: See Section~\ref{sec:conclusion}.
    \item[] Guidelines:
    \begin{itemize}
        \item The answer \answerNA{} means that the paper has no limitation while the answer \answerNo{} means that the paper has limitations, but those are not discussed in the paper. 
        \item The authors are encouraged to create a separate ``Limitations'' section in their paper.
        \item The paper should point out any strong assumptions and how robust the results are to violations of these assumptions (e.g., independence assumptions, noiseless settings, model well-specification, asymptotic approximations only holding locally). The authors should reflect on how these assumptions might be violated in practice and what the implications would be.
        \item The authors should reflect on the scope of the claims made, e.g., if the approach was only tested on a few datasets or with a few runs. In general, empirical results often depend on implicit assumptions, which should be articulated.
        \item The authors should reflect on the factors that influence the performance of the approach. For example, a facial recognition algorithm may perform poorly when image resolution is low or images are taken in low lighting. Or a speech-to-text system might not be used reliably to provide closed captions for online lectures because it fails to handle technical jargon.
        \item The authors should discuss the computational efficiency of the proposed algorithms and how they scale with dataset size.
        \item If applicable, the authors should discuss possible limitations of their approach to address problems of privacy and fairness.
        \item While the authors might fear that complete honesty about limitations might be used by reviewers as grounds for rejection, a worse outcome might be that reviewers discover limitations that aren't acknowledged in the paper. The authors should use their best judgment and recognize that individual actions in favor of transparency play an important role in developing norms that preserve the integrity of the community. Reviewers will be specifically instructed to not penalize honesty concerning limitations.
    \end{itemize}

\item {\bf Theory assumptions and proofs}
    \item[] Question: For each theoretical result, does the paper provide the full set of assumptions and a complete (and correct) proof?
    \item[] Answer: \answerNA{} 
    \item[] Justification: The paper does not include theoretical results.
    \item[] Guidelines:
    \begin{itemize}
        \item The answer \answerNA{} means that the paper does not include theoretical results. 
        \item All the theorems, formulas, and proofs in the paper should be numbered and cross-referenced.
        \item All assumptions should be clearly stated or referenced in the statement of any theorems.
        \item The proofs can either appear in the main paper or the supplemental material, but if they appear in the supplemental material, the authors are encouraged to provide a short proof sketch to provide intuition. 
        \item Inversely, any informal proof provided in the core of the paper should be complemented by formal proofs provided in appendix or supplemental material.
        \item Theorems and Lemmas that the proof relies upon should be properly referenced. 
    \end{itemize}

    \item {\bf Experimental result reproducibility}
    \item[] Question: Does the paper fully disclose all the information needed to reproduce the main experimental results of the paper to the extent that it affects the main claims and/or conclusions of the paper (regardless of whether the code and data are provided or not)?
    \item[] Answer: \answerYes{} 
    \item[] Justification: See Section~\ref{sec:surrogate_benchmarks},~\ref{sec:scan_methodology_evaluation} and Appendix~\ref{app:foundation_model_pipeline_details},~\ref{app:surrogate_modeling_details},~\ref{app:surrogate_api}. 
    \item[] Guidelines:
    \begin{itemize}
        \item The answer \answerNA{} means that the paper does not include experiments.
        \item If the paper includes experiments, a \answerNo{} answer to this question will not be perceived well by the reviewers: Making the paper reproducible is important, regardless of whether the code and data are provided or not.
        \item If the contribution is a dataset and\slash or model, the authors should describe the steps taken to make their results reproducible or verifiable. 
        \item Depending on the contribution, reproducibility can be accomplished in various ways. For example, if the contribution is a novel architecture, describing the architecture fully might suffice, or if the contribution is a specific model and empirical evaluation, it may be necessary to either make it possible for others to replicate the model with the same dataset, or provide access to the model. In general. releasing code and data is often one good way to accomplish this, but reproducibility can also be provided via detailed instructions for how to replicate the results, access to a hosted model (e.g., in the case of a large language model), releasing of a model checkpoint, or other means that are appropriate to the research performed.
        \item While NeurIPS does not require releasing code, the conference does require all submissions to provide some reasonable avenue for reproducibility, which may depend on the nature of the contribution. For example
        \begin{enumerate}
            \item If the contribution is primarily a new algorithm, the paper should make it clear how to reproduce that algorithm.
            \item If the contribution is primarily a new model architecture, the paper should describe the architecture clearly and fully.
            \item If the contribution is a new model (e.g., a large language model), then there should either be a way to access this model for reproducing the results or a way to reproduce the model (e.g., with an open-source dataset or instructions for how to construct the dataset).
            \item We recognize that reproducibility may be tricky in some cases, in which case authors are welcome to describe the particular way they provide for reproducibility. In the case of closed-source models, it may be that access to the model is limited in some way (e.g., to registered users), but it should be possible for other researchers to have some path to reproducing or verifying the results.
        \end{enumerate}
    \end{itemize}

\item {\bf Open access to data and code}
    \item[] Question: Does the paper provide open access to the data and code, with sufficient instructions to faithfully reproduce the main experimental results, as described in supplemental material?
    \item[] Answer: \answerYes{} 
    \item[] Justification: The paper does provide open access to the data and code, which are in synchronization with supplemental material.
    \item[] Guidelines:
    \begin{itemize}
        \item The answer \answerNA{} means that paper does not include experiments requiring code.
        \item Please see the NeurIPS code and data submission guidelines (\url{https://neurips.cc/public/guides/CodeSubmissionPolicy}) for more details.
        \item While we encourage the release of code and data, we understand that this might not be possible, so \answerNo{} is an acceptable answer. Papers cannot be rejected simply for not including code, unless this is central to the contribution (e.g., for a new open-source benchmark).
        \item The instructions should contain the exact command and environment needed to run to reproduce the results. See the NeurIPS code and data submission guidelines (\url{https://neurips.cc/public/guides/CodeSubmissionPolicy}) for more details.
        \item The authors should provide instructions on data access and preparation, including how to access the raw data, preprocessed data, intermediate data, and generated data, etc.
        \item The authors should provide scripts to reproduce all experimental results for the new proposed method and baselines. If only a subset of experiments are reproducible, they should state which ones are omitted from the script and why.
        \item At submission time, to preserve anonymity, the authors should release anonymized versions (if applicable).
        \item Providing as much information as possible in supplemental material (appended to the paper) is recommended, but including URLs to data and code is permitted.
    \end{itemize}

\item {\bf Experimental setting/details}
    \item[] Question: Does the paper specify all the training and test details (e.g., data splits, hyperparameters, how they were chosen, type of optimizer) necessary to understand the results?
    \item[] Answer: \answerYes{} 
    \item[] Justification: See Section ~\ref{sec:surrogate_benchmarks} and Appendix ~\ref{app:foundation_model_pipeline_details} and ~\ref{app:surrogate_modeling_details}.
    \item[] Guidelines:
    \begin{itemize}
        \item The answer \answerNA{} means that the paper does not include experiments.
        \item The experimental setting should be presented in the core of the paper to a level of detail that is necessary to appreciate the results and make sense of them.
        \item The full details can be provided either with the code, in appendix, or as supplemental material.
    \end{itemize}

\item {\bf Experiment statistical significance}
    \item[] Question: Does the paper report error bars suitably and correctly defined or other appropriate information about the statistical significance of the experiments?
    \item[] Answer: \answerYes{} 
    \item[] Justification: Standard error of the mean is shown in figures in Section~\ref{sec:results}, and in Appendix~\ref{app:scan_method_design_choices}.
    \item[] Guidelines:
    \begin{itemize}
        \item The answer \answerNA{} means that the paper does not include experiments.
        \item The authors should answer \answerYes{} if the results are accompanied by error bars, confidence intervals, or statistical significance tests, at least for the experiments that support the main claims of the paper.
        \item The factors of variability that the error bars are capturing should be clearly stated (for example, train/test split, initialization, random drawing of some parameter, or overall run with given experimental conditions).
        \item The method for calculating the error bars should be explained (closed form formula, call to a library function, bootstrap, etc.)
        \item The assumptions made should be given (e.g., Normally distributed errors).
        \item It should be clear whether the error bar is the standard deviation or the standard error of the mean.
        \item It is OK to report 1-sigma error bars, but one should state it. The authors should preferably report a 2-sigma error bar than state that they have a 96\% CI, if the hypothesis of Normality of errors is not verified.
        \item For asymmetric distributions, the authors should be careful not to show in tables or figures symmetric error bars that would yield results that are out of range (e.g., negative error rates).
        \item If error bars are reported in tables or plots, the authors should explain in the text how they were calculated and reference the corresponding figures or tables in the text.
    \end{itemize}

\item {\bf Experiments compute resources}
    \item[] Question: For each experiment, does the paper provide sufficient information on the computer resources (type of compute workers, memory, time of execution) needed to reproduce the experiments?
    \item[] Answer: \answerYes{} 
    \item[] Justification: See Table~\ref{tab:runtime_summary} and Table~\ref{tab:total_compute_cost}.
    \item[] Guidelines:
    \begin{itemize}
        \item The answer \answerNA{} means that the paper does not include experiments.
        \item The paper should indicate the type of compute workers CPU or GPU, internal cluster, or cloud provider, including relevant memory and storage.
        \item The paper should provide the amount of compute required for each of the individual experimental runs as well as estimate the total compute. 
        \item The paper should disclose whether the full research project required more compute than the experiments reported in the paper (e.g., preliminary or failed experiments that didn't make it into the paper). 
    \end{itemize}
    
\item {\bf Code of ethics}
    \item[] Question: Does the research conducted in the paper conform, in every respect, with the NeurIPS Code of Ethics \url{https://neurips.cc/public/EthicsGuidelines}?
    \item[] Answer: \answerYes{} 
    \item[] Justification: Our research conforms to the NeurIPS Code of Ethics.
    \item[] Guidelines:
    \begin{itemize}
        \item The answer \answerNA{} means that the authors have not reviewed the NeurIPS Code of Ethics.
        \item If the authors answer \answerNo, they should explain the special circumstances that require a deviation from the Code of Ethics.
        \item The authors should make sure to preserve anonymity (e.g., if there is a special consideration due to laws or regulations in their jurisdiction).
    \end{itemize}

\item {\bf Broader impacts}
    \item[] Question: Does the paper discuss both potential positive societal impacts and negative societal impacts of the work performed?
    \item[] Answer: \answerYes{} 
    \item[] Justification: See Appendix~\ref{app:broader_impacts}.
    \item[] Guidelines:
    \begin{itemize}
        \item The answer \answerNA{} means that there is no societal impact of the work performed.
        \item If the authors answer \answerNA{} or \answerNo, they should explain why their work has no societal impact or why the paper does not address societal impact.
        \item Examples of negative societal impacts include potential malicious or unintended uses (e.g., disinformation, generating fake profiles, surveillance), fairness considerations (e.g., deployment of technologies that could make decisions that unfairly impact specific groups), privacy considerations, and security considerations.
        \item The conference expects that many papers will be foundational research and not tied to particular applications, let alone deployments. However, if there is a direct path to any negative applications, the authors should point it out. For example, it is legitimate to point out that an improvement in the quality of generative models could be used to generate Deepfakes for disinformation. On the other hand, it is not needed to point out that a generic algorithm for optimizing neural networks could enable people to train models that generate Deepfakes faster.
        \item The authors should consider possible harms that could arise when the technology is being used as intended and functioning correctly, harms that could arise when the technology is being used as intended but gives incorrect results, and harms following from (intentional or unintentional) misuse of the technology.
        \item If there are negative societal impacts, the authors could also discuss possible mitigation strategies (e.g., gated release of models, providing defenses in addition to attacks, mechanisms for monitoring misuse, mechanisms to monitor how a system learns from feedback over time, improving the efficiency and accessibility of ML).
    \end{itemize}
    
\item {\bf Safeguards}
    \item[] Question: Does the paper describe safeguards that have been put in place for responsible release of data or models that have a high risk for misuse (e.g., pre-trained language models, image generators, or scraped datasets)?
    \item[] Answer: \answerNA{} 
    \item[] Justification:
    \item[] Guidelines:
    \begin{itemize}
        \item The answer \answerNA{} means that the paper poses no such risks.
        \item Released models that have a high risk for misuse or dual-use should be released with necessary safeguards to allow for controlled use of the model, for example by requiring that users adhere to usage guidelines or restrictions to access the model or implementing safety filters. 
        \item Datasets that have been scraped from the Internet could pose safety risks. The authors should describe how they avoided releasing unsafe images.
        \item We recognize that providing effective safeguards is challenging, and many papers do not require this, but we encourage authors to take this into account and make a best faith effort.
    \end{itemize}

\item {\bf Licenses for existing assets}
    \item[] Question: Are the creators or original owners of assets (e.g., code, data, models), used in the paper, properly credited and are the license and terms of use explicitly mentioned and properly respected?
    \item[] Answer: \answerYes{} 
    \item[] Justification: Existing datasets, codebases, and evaluation suites used in this work are documented in the released repositories, including references to their respective licenses and terms of use where applicable.
    \item[] Guidelines:
    \begin{itemize}
        \item The answer \answerNA{} means that the paper does not use existing assets.
        \item The authors should cite the original paper that produced the code package or dataset.
        \item The authors should state which version of the asset is used and, if possible, include a URL.
        \item The name of the license (e.g., CC-BY 4.0) should be included for each asset.
        \item For scraped data from a particular source (e.g., website), the copyright and terms of service of that source should be provided.
        \item If assets are released, the license, copyright information, and terms of use in the package should be provided. For popular datasets, \url{paperswithcode.com/datasets} has curated licenses for some datasets. Their licensing guide can help determine the license of a dataset.
        \item For existing datasets that are re-packaged, both the original license and the license of the derived asset (if it has changed) should be provided.
        \item If this information is not available online, the authors are encouraged to reach out to the asset's creators.
    \end{itemize}

\item {\bf New assets}
    \item[] Question: Are new assets introduced in the paper well documented and is the documentation provided alongside the assets?
    \item[] Answer: \answerYes{} 
    \item[] Justification: Documentation and implementation details for the released benchmark assets are provided in the repository:~\url{https://github.com/automl/scan_bench/tree/neurips2026}.
    \item[] Guidelines:
    \begin{itemize}
        \item The answer \answerNA{} means that the paper does not release new assets.
        \item Researchers should communicate the details of the dataset\slash code\slash model as part of their submissions via structured templates. This includes details about training, license, limitations, etc. 
        \item The paper should discuss whether and how consent was obtained from people whose asset is used.
        \item At submission time, remember to anonymize your assets (if applicable). You can either create an anonymized URL or include an anonymized zip file.
    \end{itemize}

\item {\bf Crowdsourcing and research with human subjects}
    \item[] Question: For crowdsourcing experiments and research with human subjects, does the paper include the full text of instructions given to participants and screenshots, if applicable, as well as details about compensation (if any)? 
    \item[] Answer: \answerNA{} 
    \item[] Justification:
    \item[] Guidelines:
    \begin{itemize}
        \item The answer \answerNA{} means that the paper does not involve crowdsourcing nor research with human subjects.
        \item Including this information in the supplemental material is fine, but if the main contribution of the paper involves human subjects, then as much detail as possible should be included in the main paper. 
        \item According to the NeurIPS Code of Ethics, workers involved in data collection, curation, or other labor should be paid at least the minimum wage in the country of the data collector. 
    \end{itemize}

\item {\bf Institutional review board (IRB) approvals or equivalent for research with human subjects}
    \item[] Question: Does the paper describe potential risks incurred by study participants, whether such risks were disclosed to the subjects, and whether Institutional Review Board (IRB) approvals (or an equivalent approval/review based on the requirements of your country or institution) were obtained?
    \item[] Answer: \answerNA{} 
    \item[] Justification:
    \item[] Guidelines:
    \begin{itemize}
        \item The answer \answerNA{} means that the paper does not involve crowdsourcing nor research with human subjects.
        \item Depending on the country in which research is conducted, IRB approval (or equivalent) may be required for any human subjects research. If you obtained IRB approval, you should clearly state this in the paper. 
        \item We recognize that the procedures for this may vary significantly between institutions and locations, and we expect authors to adhere to the NeurIPS Code of Ethics and the guidelines for their institution. 
        \item For initial submissions, do not include any information that would break anonymity (if applicable), such as the institution conducting the review.
    \end{itemize}

\item {\bf Declaration of LLM usage}
    \item[] Question: Does the paper describe the usage of LLMs if it is an important, original, or non-standard component of the core methods in this research? Note that if the LLM is used only for writing, editing, or formatting purposes and does \emph{not} impact the core methodology, scientific rigor, or originality of the research, declaration is not required.
    \item[] Answer: \answerNA{} 
    \item[] Justification: The proposed methodology does not use LLMs as a core methodological component. They are only one of the benchmarked model families.
    \item[] Guidelines:
    \begin{itemize}
        \item The answer \answerNA{} means that the core method development in this research does not involve LLMs as any important, original, or non-standard components.
        \item Please refer to our LLM policy in the NeurIPS handbook for what should or should not be described.
    \end{itemize}

\end{enumerate}

\newpage
\appendix
\section{Foundation model pipeline details}
\label{app:foundation_model_pipeline_details}

\subsection{ScAn spaces}
\label{app:scan_space}
The ScAn search space is designed to jointly explore both architectural scale and training hyperparameters. Below we explain the search space for each family.

\subsubsection{LLM}
Our space comprises nine dimensions organized into two groups: hyperparameters (learning rate, weight decay, $\beta_1$, $\beta_2$, cooldown fraction) and scale parameters (embedding size - $d_{model}$, number of layers, number of heads, token budget).

The ranges for each parameter are informed by established scaling law literature. For architectural dimensions, we follow the model size progressions studied by Hoffmann et al.~\cite{hoffmann-neurips22a}, covering models from approximately 16M to 1B parameters. The token budget range spans sub-optimal to over-trained regimes relative to compute-optimal scaling. Hyperparameters are drawn from ranges mentioned in the Step Law~\cite{li-arxiv25a}, which provides scaling relationships for batch size and learning rate. The AdamW~\cite{loshchilov-iclr19a} momentum coefficients ($\beta_1$, $\beta_2$) and weight decay bounds reflect common practice in transformer training. Full scaling and hyperparameter ranges are shown in Table~\ref{tab:llm_space_scan}. Notably, batch size is not sampled directly, instead we derive it from the token budget using the Step Law's prescribed relationship, ensuring compute-optimal throughput for each configuration, with ranges depicted in Table~\ref{tab:llm_batch_size_token_budget}.

\begin{table}[ht]
\centering
\caption{LLM ScAn Space Configuration}
\label{tab:llm_space_scan}
\begin{tabular}{lccc}
\toprule
Parameters & Type & Range / Set & Sampling \\ \midrule
 Learning rate & Float & $[1 \times 10^{-5}, 1 \times 10^{-2}]$ & Log \\
Weight decay & Float & $[1 \times 10^{-3}, 0.2]$ & Log \\
$\beta_1$ & Float & $[0.8, 0.99]$ & Log \\
$\beta_2$ & Float & $[0.9, 0.999]$ & Log \\
Cooldown fraction & Float & $[0.0, 0.3]$ & Linear \\ \midrule
Embedding size ($d_{model}$) & Categorical & $\{256, 320, \dots, 1600\}$ & Linear \\
 Number of layers & Categorical & $\{4, 5, \dots, 30\}$ & Linear \\
Number of heads & Categorical & $\{6, 8, \dots, 40\}$ & Linear \\
Number of tokens & Float & $[2 \times 10^{8}, 4 \times 10^{10}]$ & Log \\ \bottomrule
\end{tabular}
\end{table}

\begin{table}[h]
\caption{Batch size heuristic ranges derived from Step Law.}
\label{tab:llm_batch_size_token_budget}
\centering
\small
\begin{tabular}{c c}
\toprule
Token Budget & Global Batch Size \\
\midrule
$[0, 10^{8})$ & 32768 \\
$[10^{8}, 10^{9})$   & 65536 \\
$[10^{9}, 10^{10})$   & 131072 \\
$[10^{10}, 2\times10^{10})$ & 262144 \\
$[2\times10^{10}, 4\times10^{10})$ & 524288 \\
\bottomrule
\end{tabular}
\end{table}

\subsubsection{VLM}

Our search space comprises three scale parameters: vision width, text width, and the number of training samples, along with six optimization hyperparameters: learning rate, weight decay, optimizer parameters ($\beta_1$, $\beta_2$, $\epsilon$), and the warmup fraction. Table~\ref{tab:vlm_scan} summarizes the full space and corresponding ranges (note that training samples is in millions), while the sampling strategy is described in the next section. The parameter bounds are guided by prior work~\cite{radford-arxiv21a, cherti-cvpr23a, nezhurina-neurips2025a}.

\begin{table}[ht]
\centering
\caption{VLM ScAn Space Configuration}
\label{tab:vlm_scan}
\begin{tabular}{lccc}
\toprule
Parameters & Type & Range / Set & Sampling \\ \midrule
Learning rate & Float & $[1\times10^{-5}, 5\times10^{-2}]$ & Log \\
Weight decay & Float & $[1\times10^{-6}, 2\times10^{-1}]$ & Log \\
Warmup fraction & Float & $[0.0, 0.75]$ & Linear \\
$\beta_1$ & Float & $[0.9, 0.99]$ & Log \\
$\beta_2$ & Float & $[0.95, 0.999]$ & Log \\
$\epsilon$ & Float & $[1\times10^{-8}, 1\times10^{-6}]$ & Log \\ \midrule
Training samples (M) & Float & $[0.6, 120.0]$ & Warped log ($\alpha=0.5$) \\
Vision width & Categorical & $\{32, 64, \dots, 512\}$ & Weighted categorical ($\alpha=0.5$) \\
Text width & Categorical & $\{32, 64, \dots, 512\}$ & Weighted categorical ($\alpha=0.5$) \\ \bottomrule
\end{tabular}
\end{table}

In addition to the search dimensions, several design choices are fixed across all configurations, and results are therefore conditioned on these settings. Table~\ref{tab:vlm_fixed} summarizes the fixed choices. All remaining parameters not listed in the table follow defaults from~\cite{cherti-cvpr23a}, such as vocabulary size and text context length. The attention head dimension (marked with * in the table) is set to 64, except for the smallest models with embedding size 32, where the attention head dimension is reduced to 32. The number of attention heads is computed as the embedding size of the respective tower divided by 64, with a minimum of one head for the smallest configurations. This applies to both the vision and text towers.

\begin{table}[h]
\caption{Fixed design choices across all VLM configurations.}
\centering
\small
\begin{tabular}{l c}
\toprule
Component & Choice \\
\midrule
Optimizer & AdamW \\
Vision backbone & Vision Transformer (ViT) \\
Image patch size & 32 \\
Precision & Mixed precision \\
Layers (vision/text) & 12 / 12 \\
Shared embedding dimension(joint space) & 384 \\
Attention head dimension & 64$^{*}$ \\
Learning rate schedule & Cosine \\
\bottomrule
\end{tabular}
\label{tab:vlm_fixed}
\end{table}

The batch size is not fixed. Instead, we derive a heuristic inspired by~\cite{nezhurina-neurips2025a} on their results on Re-LAION-1.4B ~\cite{laion-relaion2024} dataset. We analyze Pareto-optimal configurations under a cosine learning rate schedule and observe that global batch size increases consistently with training compute. Based on this observation, we construct a mapping from the number of training samples to the corresponding global batch size. Table~\ref{tab:batch_heuristic} summarizes the resulting heuristic. An alternative mapping from FLOPs to batch size was considered, but it leads to undesirable regimes in our setting with smaller models. In particular, very small models trained on a large number of samples would receive extremely small batch sizes, resulting in inefficient and excessively long training times.

\begin{table}[h]
\caption{Batch size heuristic derived from Pareto-optimal OpenCLIP configurations.}
\centering
\small
\begin{tabular}{c c}
\toprule
Training Samples & Global Batch Size \\
\midrule
$[0, 3.4\text{M})$   & 512 \\
$[3.4\text{M}, 14.6\text{M})$   & 2,048 \\
$[14.6\text{M}, 70.4\text{M})$   & 4,096 \\
$[70.4\text{M}, 145.92\text{M})$   & 8,192 \\
$[145.92\text{M}, 340.48\text{M})$   & 16,384 \\
\bottomrule
\end{tabular}
\label{tab:batch_heuristic}
\end{table}

\subsection{Sampling strategies}
\label{app:surrogate_sampling_strategies}
Across all model families, most parameters are sampled via random search, with continuous parameters drawn either uniformly or log-uniformly depending on their scale. In addition, some parameters follow custom sampling strategies, which are described in detail in the following sections. The search space tables \ref{tab:llm_space_scan} and \ref{tab:vlm_scan} both include sampling columns that provide a brief overview of these strategies.

\subsubsection{LLM}

Configurations are sampled from the search space defined in Table~\ref{tab:llm_space_scan} following the general procedure described above. Each sampled configuration fully specifies the model architecture (embedding size, depth, number of heads), the training budget (token count), and the hyperparameters (learning rate, weight decay, $\beta_1$, $\beta_2$, and cooldown fraction).

\subsubsection{VLM}

The hyperparameters are sampled according to the general procedure described in Table~\ref{tab:vlm_scan}, with additional specific strategies detailed in the following.

The number of training samples \(N\) is drawn from a warped log-uniform distribution:
    \begin{equation}
    N = N_{\min} \left( \frac{N_{\max}}{N_{\min}} \right)^{u^{\alpha}}, 
    \quad u \sim \mathcal{U}(0,1),
    \end{equation}
    where \(N_{\min} = 6 \times 10^5\), \(N_{\max} = 1.2 \times 10^8\), and \(\alpha = 0.5\).

    A standard log-uniform distribution would allocate equal probability mass to each order of magnitude, resulting in half of the configurations being sampled below \(6 \times 10^6\) samples. This was found to be overly restrictive. The warped formulation (\(\alpha < 1\)) provides a controlled trade-off by relaxing the equal-per-decade constraint of log-uniform sampling.

Architectural parameters, namely vision and text width, are selected from the discrete set \(\{32, 64, 128, 192, 256, 320, 384, 448, 512\}\). To bias sampling toward smaller model sizes, we assign a weight to each candidate based on its index. Let \(i \in \{1, \dots, K\}\) denote the index of a candidate (ordered from smallest to largest). The weights are defined as
    \begin{equation}
    w_i = \frac{1}{i^{\alpha}},
    \end{equation}
    with \(\alpha = 0.5\), such that smaller widths receive higher weights. These weights are normalized to obtain a probability distribution:
    \begin{equation}
    P(i) = \frac{w_i}{\sum_{j=1}^{K} w_j},
    \end{equation}
    from which the final width is sampled.

\subsection{Multi-fidelity modeling and checkpointing}
\label{app:multi_fidelity_modeling_checkpoints}

We record intermediate checkpoints for each configuration to enable multi-fidelity modeling in the surrogate benchmarks. Each checkpoint corresponds to a different stage of training. To capture this, we define a fidelity variable \texttt{training\_progress} $\in [0,1]$, representing normalized training progress, where 0 corresponds to initialization and 1 to the fully trained model.
Each checkpoint is mapped to a value of \texttt{training\_progress} based on its relative position within the training process. This yields multiple observations per configuration at different fidelity levels, enabling the surrogate to model performance as a function of both configuration and training progress.

\subsubsection{LLM}
We determine the number of training steps for each model based on a fixed token budget. Specifically,
\begin{equation}
\label{eq:number_steps}
N_{\text{steps}} = \left\lfloor \frac{T}{S \cdot B_{\text{global}}} \right\rfloor,
\end{equation}
where $T$ denotes the target token budget, $S = 2{,}048$ is the sequence length, and 
\begin{equation}
B_{\text{global}} = \texttt{micro\_batch\_size} \times \texttt{gradient\_accumulation\_steps} \times \texttt{num\_gpus}
\end{equation}
is the global batch size.

Let $N_{\text{steps}}^{(m)}$ denote the number of steps for model $m$, and let
\begin{equation}
N_{\text{steps}}^{\max} = \max_m N_{\text{steps}}^{(m)}
\end{equation}
be the maximum number of steps across all models.

The number of checkpoints assigned to each model is then given by:
\begin{equation}
N_{\text{ckpt}}^{(m)} = \max\left(2,\ \left\lceil 10 \cdot \frac{N_{\text{steps}}^{(m)}}{N_{\text{steps}}^{\max}} \right\rceil \right).
\end{equation}

Thus, the model with the largest number of training steps is assigned 10 checkpoints, while all other models receive between 2 and 10 checkpoints, proportional to their number of training steps.

\subsubsection{VLM}
The number of epochs, which determines the number of recorded checkpoints, is computed as:
\begin{equation}
E = \max\left(2,\ \left\lceil \frac{N_{\text{planned}}}{N_{\text{max per epoch}}} \right\rceil \right),
\end{equation}
where $N_{\text{planned}}$ denotes the planned training budget, defined as the number of training samples. The constant $N_{\text{max per epoch}}$ controls the frequency of checkpointing, and is set to 6 million. For each configuration we have 2 to 10 checkpoints.

\subsection{Model training and evaluation}

\label{app:model_training_and_evaluation}

\subsubsection{LLM}
\paragraph{Training}
Every configuration is trained from scratch on a designated subset of the SlimPajama~\cite{soboleva-cerebras2023a} dataset using a decoder-only transformer with AdamW~\cite{loshchilov-iclr19a} as an optimizer, SwiGLU~\cite{shazeer-arxiv2020a} activations, RMSNorm~\cite{zhang-neurips19b}, weight tying, a vocabulary size of 50\,277 and a sequence length of 2\,048. Training follows a warmup-stable-decay (WSD)~\cite{hu-arxiv24a} learning rate schedule. Full training information is depicted in Table~\ref{tab:llm_training_configuration}. The total number of training steps is determined by the token budget and global batch size as detailed in Equation~\eqref{eq:number_steps}.
\begin{table}[h]
\caption{Fixed training choices across all LLM configurations.}
\label{tab:llm_training_configuration}
\centering
\small
\begin{tabular}{l c}
\toprule
Component & Choice \\
\midrule
Optimizer & AdamW \\
Vocabulary size & 50277 \\
Sequence length & 2048 \\   
Activation function & SwiGLU \\
Learning rate schedule & WSD \\
Normalization & RMSNorm \\
Weight tying & True \\

\bottomrule
\end{tabular}
\end{table}

\paragraph{Evaluation}
At each evaluation step, upstream performance is measured via validation and test loss on the respective SlimPajama~\cite{soboleva-cerebras2023a} split, while downstream performance is evaluated on six benchmarks (HellaSwag~\cite{zellers-acl2019a}, PIQA~\cite{bisk-aaai2020a}, OpenBookQA~\cite{mihaylov-acl2018a}, ARC Easy~\cite{clark-arxiv2018a}, ARC Challenge~\cite{clark-arxiv2018a}, COPA~\cite{roemmele-aaai2011a}) using the respective few-shot settings.

\subsubsection{VLM}

\paragraph{Training} Models are trained using the AdamW~\cite{loshchilov-iclr19a} optimizer with a global batch size determined by the heuristic described earlier and gradient accumulation set to 1. Training is performed in a distributed setting using local loss computation instead of constructing the full global similarity matrix.

\paragraph{Evaluation} Each trained model is evaluated on separate validation and test sets to obtain performance metrics. Both splits are drawn from pre-training dataset, with 100{,}000 samples each. Evaluation is performed with a fixed batch size of 128 to ensure comparability across configurations.

For downstream evaluation, we use the DataComp evaluation suite~\cite{gadre-neurips2023a}, which comprises 40 datasets evaluated in a zero-shot setting without additional training. The suite spans multiple task categories, including classification, retrieval, robustness, and fairness benchmarks. As individual tasks report different evaluation metrics, we model the primary metric specified for each task by~\cite{gadre-neurips2023a}. For FAIRFACE~\cite{karkkainen-wacv2021a} and UTKFACE~\cite{zhang-cvpr2017a}, which do not define a single primary metric, we use \texttt{acc\_race\_avg}. 

\begin{table}[t]
\centering
\caption{Compute cost (GPU hours) for data collection across different GPU types. The total column reports the sum of training and evaluation costs across both VLM and LLM benchmarks.}
\label{tab:total_compute_cost}
\begin{tabular}{lccccccc}
\toprule
& \multicolumn{2}{c}{Training} 
& \multicolumn{2}{c}{Downstream} 
& \multicolumn{2}{c}{Upstream}
& Total \\
\cmidrule(lr){2-3} \cmidrule(lr){4-5} \cmidrule(lr){6-7}
GPU 
& VLM & LLM 
& VLM & LLM 
& VLM & LLM 
&  \\
\midrule
RTX 2080 Ti & 11.84  & - & 731.84  & - & 11.39 & - & 755.07 \\
RTX 3080  & 142.42 & - & 373.53 & - & 13.47 & - & 529.42 \\
L40     & 0.44 & - & - & - & 9.84 & - & 10.28 \\
A100 40GB    & 3986.02 & 11042.45 & 828.75 & 935.35 & 198.66 & 3972.34 & 20962.55 \\
H100 94GB    & 6653.11 & 26870.62 & 859.65 & - & 159.61 & 348.27  & 34891.26 \\
\bottomrule
\end{tabular}
\end{table}

\section{Surrogate modeling details}
\label{app:surrogate_modeling_details}

\subsection{Surrogate candidates}
\label{app:surrogate_candidates}

We adopt the surrogate modeling setup from HW-GPT-Bench \cite{sukthanker-neurips24a}, including AutoGluon, XGBoost, LightGBM, and mixed ensembles, as described in Appendix C.2 of the original work.

We introduce three modifications. First, we exclude the XGBoost learning rate $1.0$, as it did not improve predictive performance in our setting. Second, we use AutoGluon v1.5 with default configurations instead of manually specified hyperparameters, as we observed no meaningful differences in accuracy. For AutoGluon, we set the training time budget to 30 minutes for LLMs and 4 hours for VLMs. Third, we exclude \texttt{FASTAI} models in AutoGluon due to training stability issues.

In addition, we include TabPFN~\cite{grinsztajn-arxiv2026a} v2.5 as an additional surrogate candidate. TabPFN has demonstrated strong performance on low-data tabular tasks, which aligns with our setting. Moreover, its in-context learning capability removes the need for explicit training, allowing efficient reuse across multiple downstream targets without training separate models for each metric.

\subsection{Data splitting}
\label{app:surrogate_data_splitting}

The dataset contains multiple checkpoints per configuration. To avoid information leakage, splitting is performed at the configuration level, assigning each configuration and all its checkpoints entirely to either the training or test set.

For VLMs and LLMs, we perform an 80/20 train test split over configurations. To ensure coverage across different compute scales, configurations are partitioned into four equally sized groups based on their training compute, and a stratified split is performed within each group. This ensures that both training and test sets include configurations from all compute ranges. 

Additionally, in VLMs we build an additional failure prediction dataset, where each configuration is assigned a binary label indicating whether it diverged during training. Configurations are randomly split into training and test sets using a 80/20 split. This dataset is defined at the configuration level, and checkpoints are not considered.

\subsection{Evaluation metrics}
\label{app:surrogate_evaluation_metrics}

We adopt the accuracy metrics from HW-GPT-Bench \cite{sukthanker-neurips24a}, including Root Mean Squared Error (RMSE), Mean Absolute Error (MAE), Median Absolute Error (MDAE), Mean Absolute Relative Percentage Deviation (MARPD), coefficient of determination ($R^2$), Pearson correlation ($R$), and Spearman rank correlation, and omit calibration metrics to focus solely on predictive performance.

Table~\ref{tab:full_surrogate_selection} summarizes surrogate performance across model families and metrics, using the respective train test split and reporting results on upstream performance for each model family.

\begin{table}[t]
\centering
\small
\caption{Surrogate performance across model families on held-out final configurations.}
\label{tab:full_surrogate_selection}

\begin{tabular}{l l ccccccc}
\toprule
Family & Surrogate 
& RMSE $\downarrow$ 
& MAE $\downarrow$ 
& MDAE $\downarrow$ 
& MARPD $\downarrow$ 
& $R^2$ $\uparrow$
& R $\uparrow$
& Corr. $\uparrow$ \\
\midrule

\multirow{5}{*}{VLM}
& TabPFN & \textbf{0.21}  & \textbf{0.06}  & \textbf{0.02} & \textbf{2.69}  & \textbf{0.96} & \textbf{0.98} & \textbf{0.98} \\
& AutoGluon & 0.26 & 0.12 & 0.07 & 5.79 & 0.94 & 0.97 & 0.98 \\
& XGB & 0.35 & 0.20 & 0.13 & 9.11 & 0.90 & 0.95 & 0.96 \\
& Mix & 0.37 & 0.21 & 0.14 & 10.11 & 0.89 & 0.95 & 0.95 \\
& LGB & 0.42 & 0.29 & 0.24 & 14.88 & 0.86 & 0.95 & 0.96 \\

\midrule

\multirow{5}{*}{LLM}
& TabPFN & $\mathbf{0.27}$
& $\mathbf{0.08}$
& $\mathbf{0.01}$
& $\mathbf{3.02}$
& $\mathbf{0.77}$
& $\mathbf{0.88}$
& $\mathbf{0.95}$ \\
& AutoGluon & $0.30$ 
& $0.11$ 
& $0.02$ 
& $4.66$ 
& $0.72$ 
& $0.85$ 
& $0.91$ \\
& XGB & $0.35$
& $0.15$
& $0.04$
& $6.80$
& $0.63$
& $0.80$
& $0.86$ \\
& Mix & $0.33$
& $0.17$
& $0.07$
& $7.65$
& $0.66 $
& $0.82 $
& $0.86 $ \\
& LGB & $0.33$
& $0.17 $
& $0.08 $
& $7.97 $
& $0.66 $
& $0.83 $
& $0.87 $ \\

\bottomrule
\end{tabular}

\end{table}

\subsection{Divergence prediction (VLMs)}
\label{app:divergence_prediction}

As described in the data splitting procedure (Appendix~\ref{app:surrogate_data_splitting}), we construct a dataset for modeling failed runs. Based on this data, we train a binary classifier that predicts whether a configuration will diverge. Empirically, divergence is primarily caused by extreme hyperparameter settings, in particular high learning rates. We therefore use a simple model based on gradient boosted decision trees. 

Specifically, we use an ensemble of XGBoost classifiers. Each model in the ensemble is an XGBoost classifier with tree-based architecture, trained with varying maximum depth $\{3,5,9\}$, number of estimators $\{300,500,800\}$, and learning rates $\{0.01,0.05,0.1\}$. 

At query time, configurations predicted to diverge are filtered out and no performance prediction is returned.

\section{Surrogate API}
\label{app:surrogate_api}

The surrogate API provides a unified interface for querying configurations at arbitrary training progress. Given a configuration and a training progress value, the API returns upstream and downstream metrics predicted by surrogate models, along with predictive uncertainty estimates and additional quantities computed from the configuration, including cumulative training compute (FLOPs) and model parameter counts. Table~\ref{tab:api_interface} provides an overview of the interface, conditioned on the configuration not being predicted to diverge in the VLM case. Configurations predicted to fail do not return performance outputs, but only auxiliary quantities such as FLOPs and parameter counts.

\subsection{Predictive uncertainty}
The API encodes predictive uncertainty via TabPFN’s quantile outputs, defined as the difference between the 90th and 10th percentiles of the predictive distribution.

\subsection{Parameter count}

\paragraph{LLMs} The total number of trainable parameters \(N\) is computed as:
\begin{equation}
N = L(4d^{2} + 3dh + 2d) + 2dV + d,
\end{equation}
where \(L\) is the number of layers, \(d\) the model embedding dimension, \(V\) the vocabulary size, and \(h\) the hidden dimension of the feed-forward network, defined as:
\begin{equation}
h = 256 \left\lceil \frac{8d/3}{256} \right\rceil.
\end{equation}

\paragraph{VLMs} Model parameter counts are computed as the total number of trainable parameters. We further report separate parameter counts for the vision encoder and the text encoder. For the text encoder, we additionally distinguish non-embedding parameters, which comprise the transformer layers, final layer normalization, and text projection layers, excluding token embeddings.

\subsection{FLOPs estimation}

\paragraph{LLMs} We approximate the total training compute \(C\) in FLOPs by accounting for both forward and backward passes. Each parameter contributes approximately 2 FLOPs per token in the forward pass, and the backward pass is estimated as twice the forward pass, yielding an overall factor of 6. The total compute is thus given by:
\begin{equation}
C = 6T \Bigl[ L \bigl( 4d^{2} + 2Sd + 3dh \bigr) + dV \Bigr],
\end{equation}
where \(T\) denotes the total number of training tokens and \(S\) the sequence length.

\paragraph{VLMs} We follow the OpenCLIP implementation for FLOPs estimation. Specifically, we use the PyTorch~\cite{paszke-neurips19a} \texttt{torch.utils.flop\_counter} profiler to measure the per-sample FLOPs of a forward pass. Total training compute is then obtained by multiplying the per-sample FLOPs by the total number of training samples.

\begin{table}[t]
\centering
\caption{Surrogate API query interface. Given a configuration and a fidelity value, the API returns performance predictions and additional metadata.}
\label{tab:api_interface}
\begin{tabular}{lll}
\toprule
Category & Field & Description \\
\midrule

\multirow{3}{*}{Input} 
& Configuration & Sampled from ScAn space~\ref{app:scan_space} \\
& Training progress & Fidelity in $[0,1]$ indicating training progress \\

\midrule

\multirow{5}{*}{Output} 
& Upstream metrics & Validation or test loss \\
& Downstream metrics & Defined in~\ref{app:model_training_and_evaluation} \\
& FLOPs & Cumulative training compute up to queried fidelity \\
& Model parameters & Number of model parameters \\
& Uncertainty & Surrogate model uncertainty \\

\bottomrule
\end{tabular}
\end{table}

\section{Broader impacts}
\label{app:broader_impacts}
Our goal is to make scaling law research for VLMs and LLMs more efficient, reproducible, and comparable. Surrogate benchmarks enable systematic evaluation of scaling methodologies without repeated large-scale training, reducing computational cost and carbon emissions. In practice, methods are often compared under different datasets, training procedures, and compute budgets, making it unclear which method performs better. Our approach enables fair and controlled comparison across techniques, while lowering the barrier to entry for large-scale model research. To the extent that this improves scaling laws, it may contribute to the development of more capable foundation models. Such progress can have both positive and negative societal impacts, as more powerful models can enable beneficial applications but may also be misused, for example to generate misleading content, amplify biases, or support large-scale surveillance. These risks arise from the underlying models rather than the benchmark itself.

\section{ScAn Methodology}
\label{app:scan_method_design_choices}
\subsection{Chinchilla-based data acquisition design choices}
We adapt generalized, heuristic-free baselines. For example, in Chinchilla acquisition approaches, rather than relying on tuned priors, we evaluate the allocation strategy across 10 random seeds, where each seed operates under a fixed $\lambda_o$ uniformly sampled from $\Lambda_o$. Furthermore, because the original work lacks a formal prescription for generating the discrete grid of evaluated models, we systematize this generation process. We construct a monotonically increasing sequence of configurations, $\lambda_{c, t_i} \le \lambda_{c, t_{i+1}}$, utilizing Sobol sampling to iteratively select which specific dimension $j$ of the scaling parameters $\lambda_c$ is augmented at each step. The detailed procedure is provided in Algorithm~\ref{alg:chinchilla_app1_single} and Algorithm~\ref{alg:chinchilla_app2_single}.

\subsection{CARBS data acquisition design choices}
We apply this same strict, prior-free constraint to the CARBS baseline. While we maintain the core acquisition behavior, the standard CARBS implementation relies on a manually specified, prior-informed search center. To eliminate this human bias, we replace it with a systematic, automated initialization protocol, ensuring a completely equitable comparison against the Chinchilla baselines. We set most of the hyperparameters to the defaults used in their original source code, providing only the search center. If a hyperparameter is in LogSpace, we use the $\sqrt{\min \cdot \max}$ value of its domain for the search center; otherwise, we compute the arithmetic mean of the boundary values to define the search center, along with defining the initial search radius.

\begin{algorithm}[htbp]
\caption{Systematized Iso-Parameter Acquisition (Single Experiment)}
\label{alg:chinchilla_app1_single}
\begin{algorithmic}[1]
\Require Random seed $k$, non-scaling parameter space $\Lambda_o$, compute budget $[C_{\text{min}}, C_{\text{max}}]$
\Require Token scaling multipliers $M$ (e.g., $\{5, 30, 60, 80\}$ for LLM)

\State Initialize experiment environment with random seed $k$
\State Initialize acquired dataset $\mathcal{D} \leftarrow \emptyset$
\State Sample fixed hyperparameters uniformly: $\lambda_o \sim \text{Uniform}(\Lambda_o)$
\State Initialize configuration sequence $\mathcal{S}_c \leftarrow \emptyset$ and base model size $\lambda_{c, t_0}$

\Comment{Phase 1: Generate monotonically increasing model size grid}
\While{grid size $<$ target horizon (e.g., 16)}
    \State Use Sobol sampling to select dimension $j$ of $\lambda_c$ to augment
    \State $\lambda_{c, t_{i+1}} \leftarrow \text{Augment}(\lambda_{c, t_i}, \text{dim}=j)$ 
    \State $\mathcal{S}_c \leftarrow \mathcal{S}_c \cup \{\lambda_{c, t_{i+1}}\}$
\EndWhile

\Comment{Phase 2: Evaluate combinations across fixed token multipliers}
\For{\textbf{each} model configuration $\lambda_c \in \mathcal{S}_c$}
    \For{\textbf{each} multiplier $m \in M$}
        \State Compute token count $D \leftarrow m \times \text{Size}(\lambda_c)$ 
        \State Estimate total compute cost $C$ for $(\lambda_c, D)$
        
        \Comment{Phase 3: Filter by compute bounds and execute}
        \If{$C_{\text{min}} \le C \le C_{\text{max}}$}
            \State $L \leftarrow \text{TrainAndEvaluate}(\lambda_c, \lambda_o, D)$
            \State $\mathcal{D} \leftarrow \mathcal{D} \cup \{(C, \lambda_c, \lambda_o, D, L)\}$
        \EndIf
    \EndFor
\EndFor

\State \textbf{return} $\mathcal{D}$
\end{algorithmic}
\end{algorithm}

\begin{algorithm}[htbp]
\caption{Systematized Iso-FLOP (Chinchilla 2) Acquisition}
\label{alg:chinchilla_app2_single}
\begin{algorithmic}[1]
\Require Random seed $k$, non-scaling parameter space $\Lambda_o$, compute bounds $C_{\text{min}}, C_{\text{max}}$
\Require Number of target FLOP steps $N_{\text{steps}}$, compute cost tolerance $\epsilon$

\State Initialize experiment environment with random seed $k$
\State Initialize acquired dataset $\mathcal{D} \leftarrow \emptyset$
\State Sample fixed hyperparameters uniformly: $\lambda_o \sim \text{Uniform}(\Lambda_o)$
\State Generate model configuration pool $\mathcal{S}_c$ via Sobol sampling

\Comment{Phase 1: Generate logarithmic compute budget grid}
\State $C_{\text{grid}} \leftarrow \text{LogSpace}(\log_{10}(2 \cdot C_{\text{min}}), \log_{10}(C_{\text{max}}), N_{\text{steps}})$

\Comment{Phase 2: Iterate over fixed target budgets and model sizes}
\For{\textbf{each} target compute budget $C_{\text{target}} \in C_{\text{grid}}$}
    \For{\textbf{each} model configuration $\lambda_c \in \mathcal{S}_c$}
        
        \Comment{Phase 3: Binary search for required token count $D$}
        \State $D_{\text{low}} \leftarrow 1$, $D_{\text{high}} \leftarrow D_{\text{limit}}$
        \State $D_{\text{match}} \leftarrow \emptyset$
        
        \While{$D_{\text{low}} \le D_{\text{high}}$}
            \State $D_{\text{mid}} \leftarrow \lfloor (D_{\text{low}} + D_{\text{high}}) / 2 \rfloor$
            \State $C_{\text{curr}} \leftarrow \text{EstimateCost}(\lambda_c, D_{\text{mid}})$
            
            \If{$|C_{\text{curr}} - C_{\text{target}}| < \epsilon$}
                \State $D_{\text{match}} \leftarrow D_{\text{mid}}$
                \State \textbf{break} \Comment{Match found within tolerance}
            \ElsIf{$C_{\text{curr}} < C_{\text{target}}$}
                \State $D_{\text{low}} \leftarrow D_{\text{mid}} + 1$
            \Else
                \State $D_{\text{high}} \leftarrow D_{\text{mid}} - 1$
            \EndIf
        \EndWhile
        
        \Comment{Phase 4: Execute if a valid configuration is resolved}
        \If{$D_{\text{match}} \neq \emptyset$}
            \State $L \leftarrow \text{TrainAndEvaluate}(\lambda_c, \lambda_o, D_{\text{match}})$
            \State $\mathcal{D} \leftarrow \mathcal{D} \cup \{(C_{\text{target}}, \lambda_c, \lambda_o, D_{\text{match}}, L)\}$
        \EndIf
        
    \EndFor
\EndFor

\State \textbf{return} $\mathcal{D}$
\end{algorithmic}
\end{algorithm}

\subsection{Loss Extrapolation }

\begin{figure*}[htbp]
    \centering
    \includegraphics[width=\textwidth]{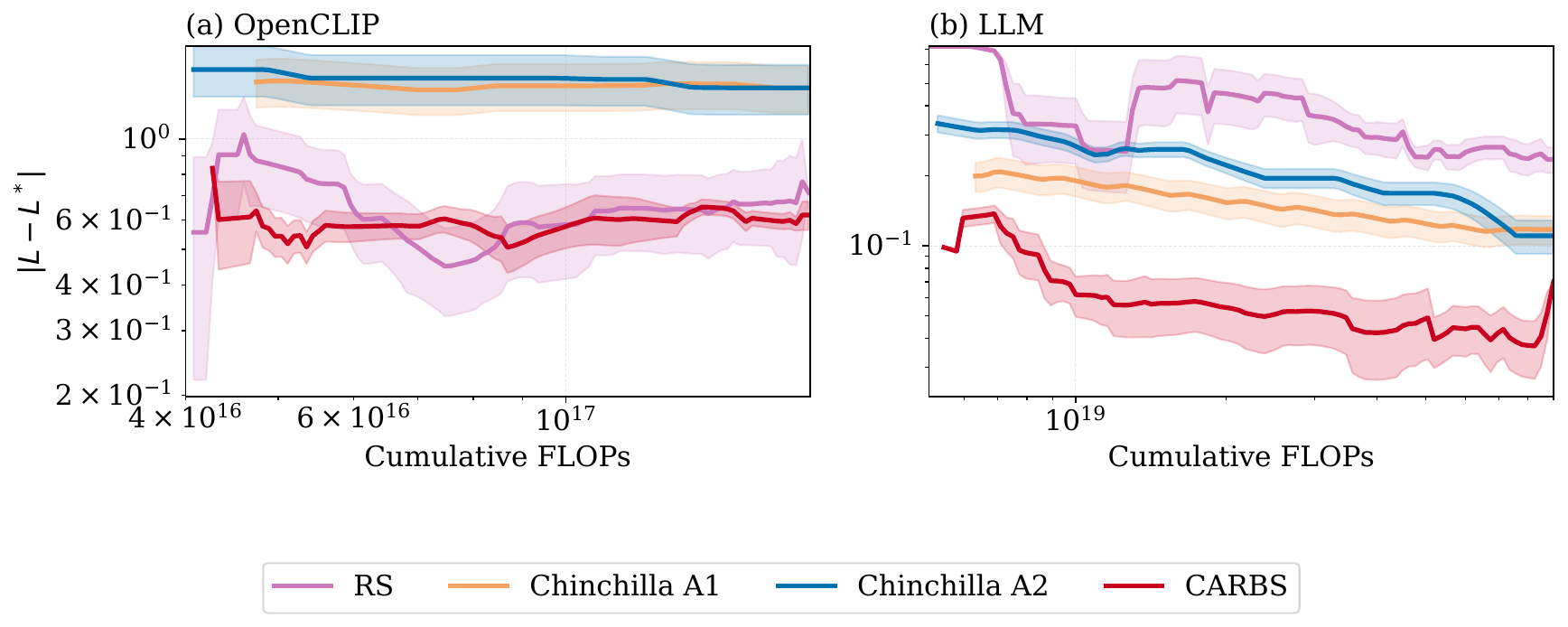}
    \caption{\textbf{Loss extrapolation regret over trajectory using Kaplan strategy}: Illustrating the loss prediction regret for all the data acquisition strategies using Kaplan extrapolation approach.}
    \label{fig:kaplan_prediction_error_trajectories}
\end{figure*}

\begin{figure*}[htbp]
    \centering
    \includegraphics[width=\textwidth]{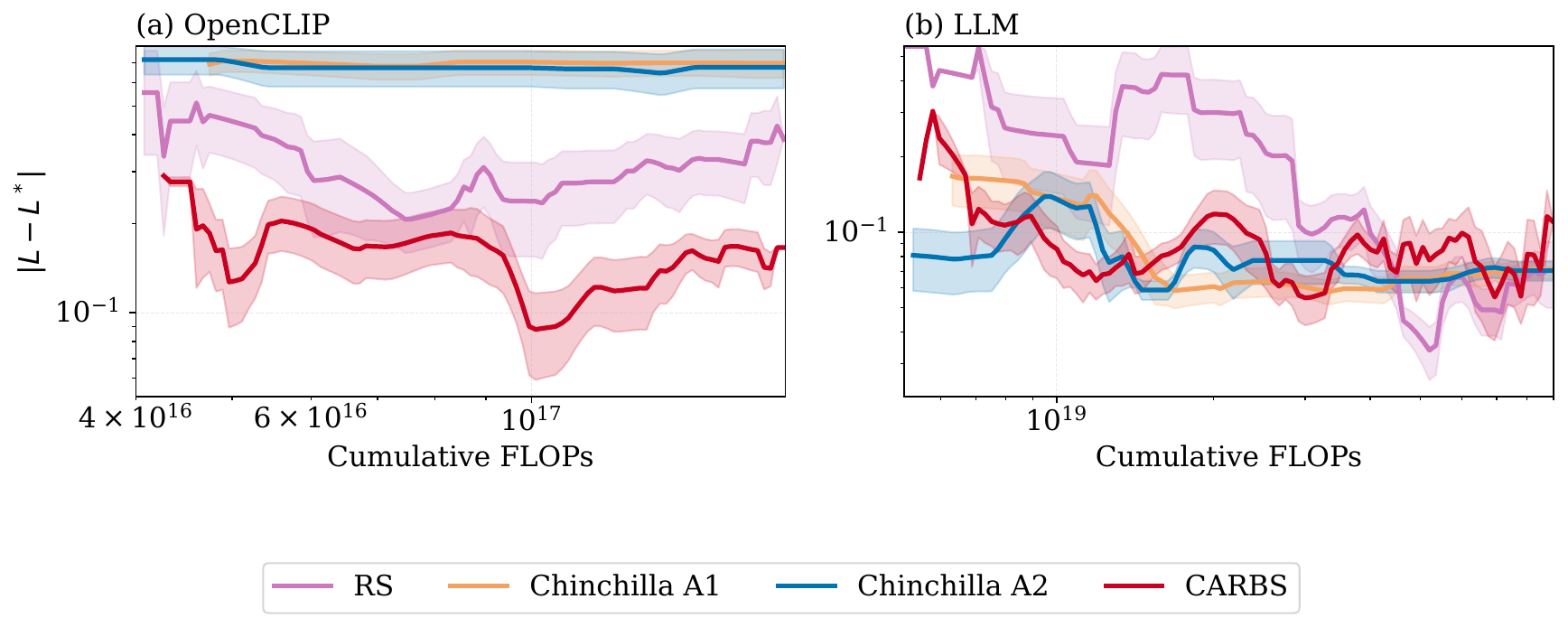}
    \caption{\textbf{Loss extrapolation regret over trajectory using Chinchilla 3 strategy}: Illustrating the loss prediction regret for all the data acquisition strategies using Chinchilla 3 extrapolation approach.}
    \label{fig:chinchilla_avg_loss}
\end{figure*}

Looking at the performance of each data acquisition strategy with the Kaplan and Chinchilla 3 approaches in Figures \ref{fig:kaplan_prediction_error_trajectories} and \ref{fig:chinchilla_avg_loss}, we show that the Chinchilla 3 approach models the optimal trajectory better in the OpenCLIP benchmark across all acquisition methods. However, the regret ranges for both approaches in the LLM benchmark are comparable, demonstrating the overall superior performance of Chinchilla 3 in modeling optimal loss.

\subsection{Coarse Extrapolation}
Here, we demonstrate the model size prediction regret of different data acquisition strategies using the Chinchilla 3 extrapolation technique, highlighting its superior performance on the LLM benchmark due to the compute cost estimation embedded within its parametric loss function.
\begin{figure*}[htbp]
    \centering
    \includegraphics[width=\textwidth]{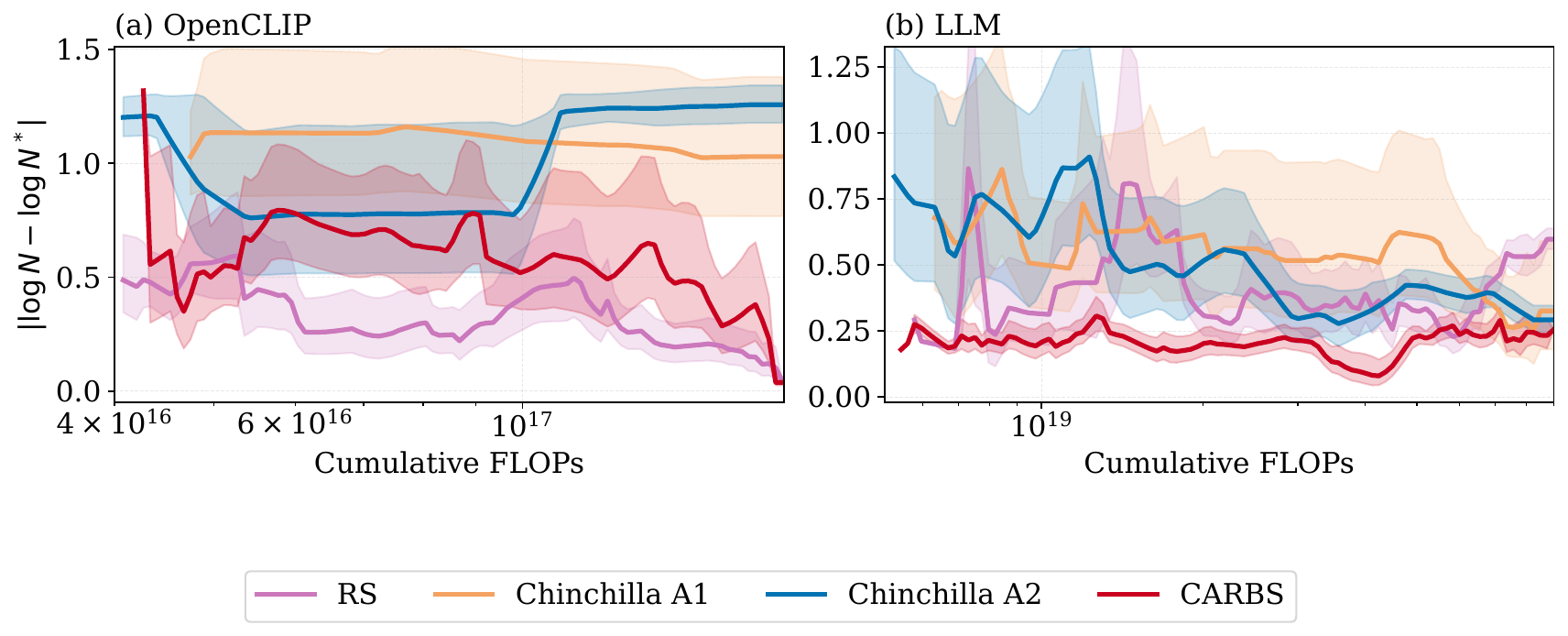}
    \caption{\textbf{Model size extrapolation regret over trajectory using the Chinchilla 3 strategy.} The plots display the absolute logarithmic regret, defined as $|\log N - \log N^*|$, which measures the deviation between the predicted optimal parameter count ($N$) and the true empirical optimum ($N^*$) as a function of cumulative profiling FLOPs. Results are shown for \textbf{(a)} the OpenCLIP and \textbf{(b)} LLM benchmarks. Shaded regions denote the standard error across independent seeds. Notably, the CARBS acquisition strategy achieves the lowest final parameter prediction regret across both modalities.}
    \label{fig:chinchilla_model_size}
\end{figure*}

\section{Ablation Studies}
\label{app:ablation}
\subsection{Robustness of Extrapolation at $10\times$ Target Scale}
\begin{figure*}[htbp]
    \centering
    \includegraphics[width=\textwidth]{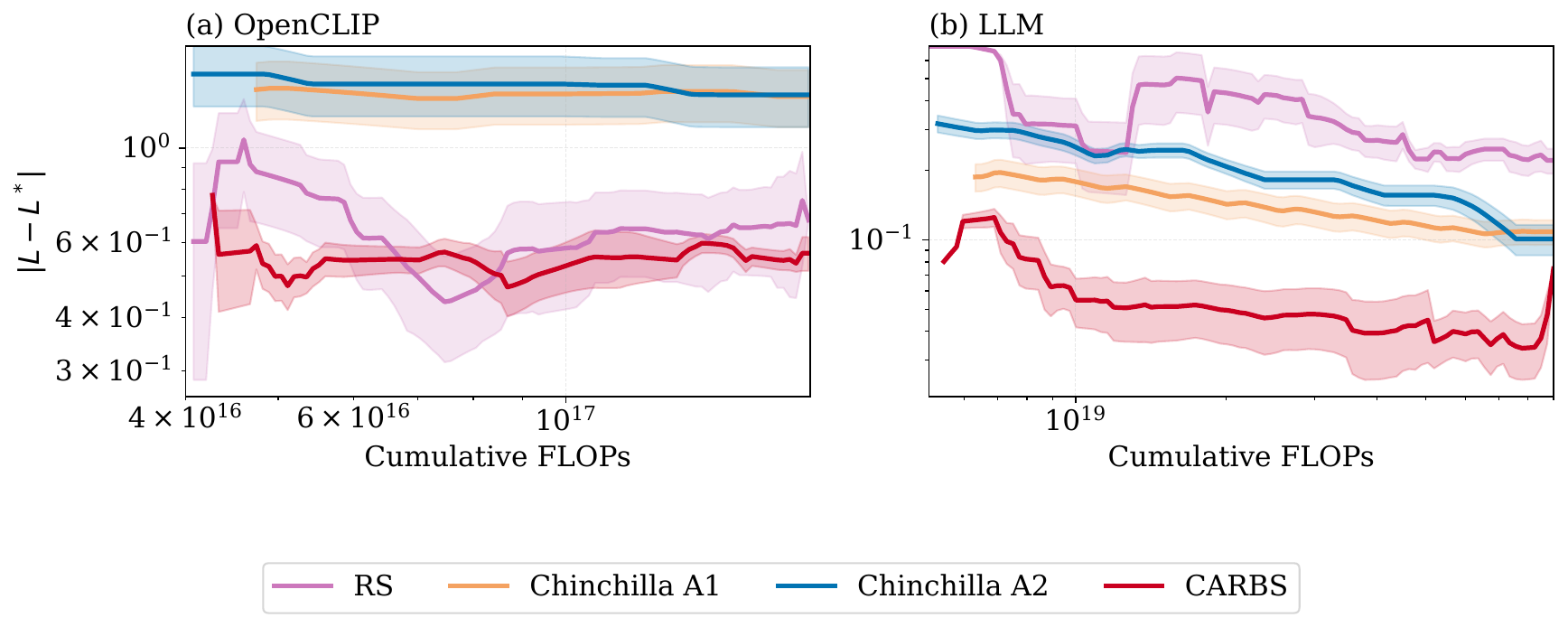}
    \caption{\textbf{Loss extrapolation trajectories at $10\times$ target scale using the Kaplan prediction strategy.} This plot illustrates the predicted loss trajectories over cumulative FLOPs across varied acquisition strategies evaluated at ($10\times C_{\text{max}}$). Consistent with baseline findings, extending the target budget induces a uniform shift in the prediction trajectory in log-log space, preserving the stable, monotonic improvement characteristic of the Kaplan method.}
    \label{fig:10x_loss_kaplan}
    \vspace{1em}
    \centering
    \includegraphics[width=\textwidth]{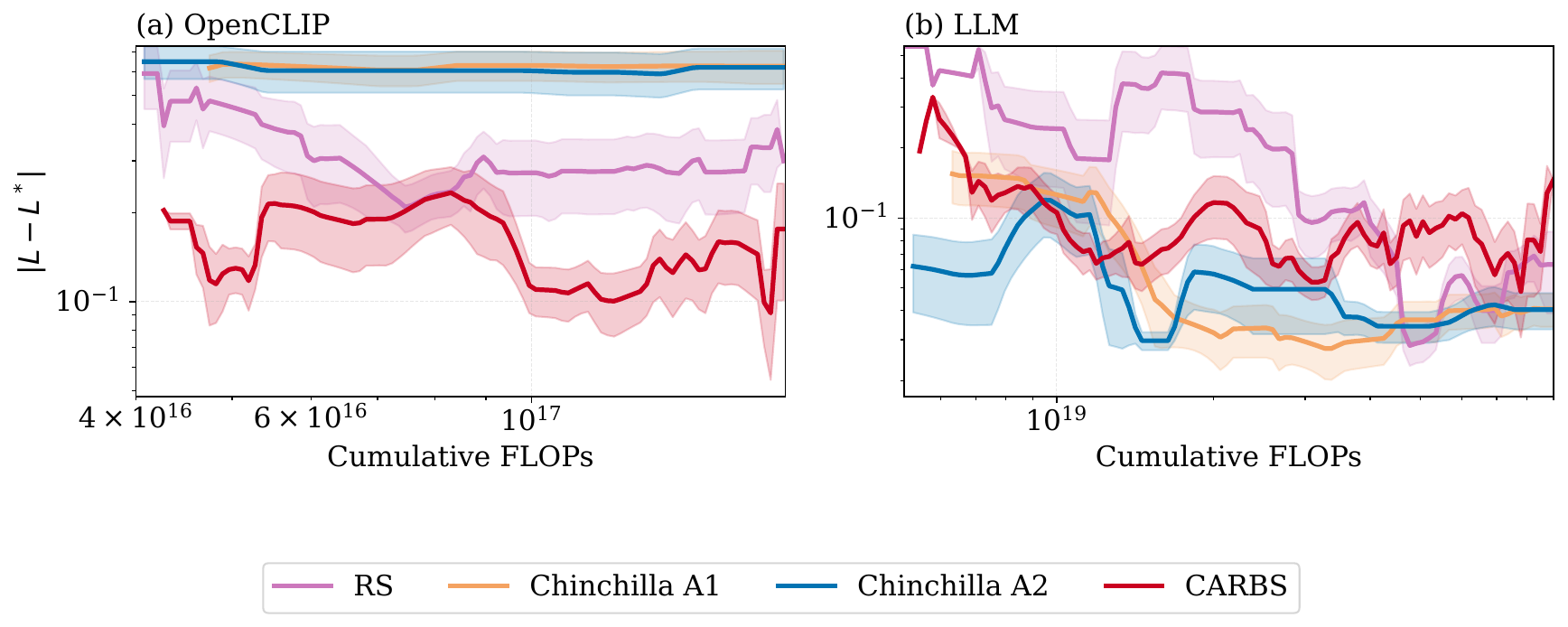}
    \caption{\textbf{Loss extrapolation trajectories at $10\times$ target scale using the Chinchilla Approach 3 strategy.} Predicted loss is evaluated at target compute budget of $10\times C_{\text{max}}$. While the large extrapolation gap naturally shifts the prediction trajectory, the comparative behavior and relative ranking of the Chinchilla Approach 3, including its stronger localized predictions on specific benchmarks, remain preserved and synchronous with the empirical loss trajectory.}
    \label{fig:10x_loss_chinchilla}

\end{figure*}

\begin{figure*}[htbp]
    
    \centering
    \includegraphics[width=\textwidth]{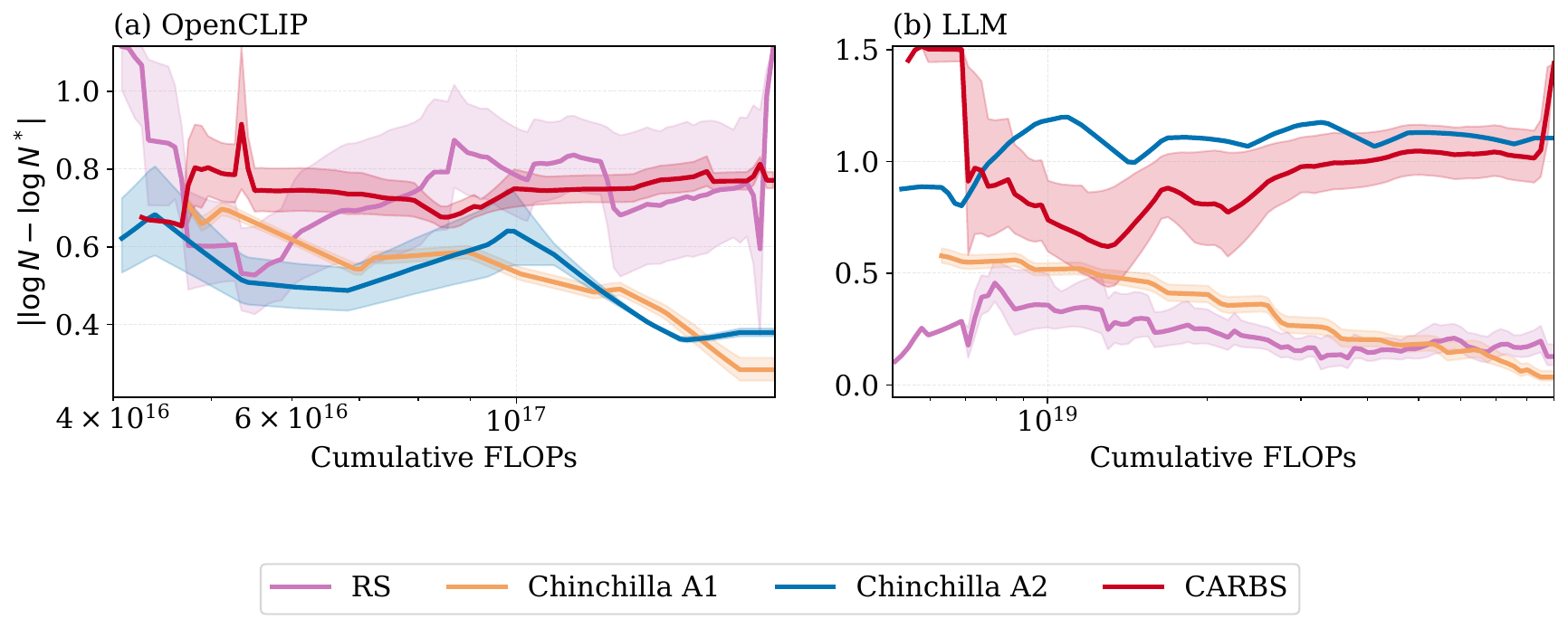}
    \caption{\textbf{Robustness of Power-Law model size ($N^*$) predictions at $10\times$ target scale.} Trajectories of the predicted optimal model size are evaluated against a $10\times$ extrapolation gap across acquisition strategies. The purely empirical Power-Law fit continues to demonstrate strong robustness across diverse architectural domains. The relative ranking of acquisition methods remains entirely consistent with baseline scale results, including the degradation of CARBS due to its localized exploitation bias.}
    \label{fig:10x_n_param_power_law}
    \vspace{1em}
    \centering
    \includegraphics[width=\textwidth]{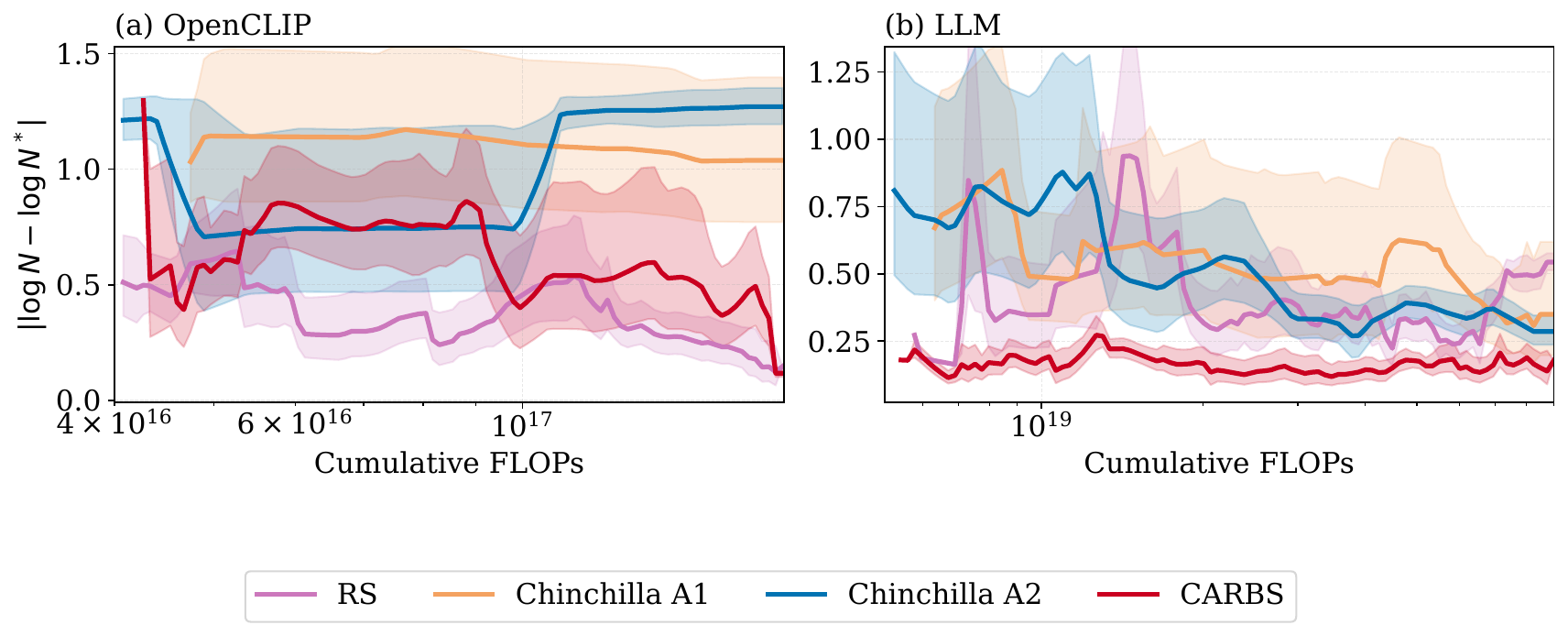}
    \caption{\textbf{Model size ($N^*$) predictions at $10\times$ target scale using the Chinchilla Approach 3 strategy.} This figure highlights the extrapolation of optimal model size. The parametric limitations of Chinchilla Approach 3 persist at this scale; specifically, its rigid reliance on standard compute estimation heuristics continues to cause structural prediction failures on multi-modal architectures like OpenCLIP.}
    \label{fig:10x_n_param_chinchilla}
\end{figure*}

We have already observed the prediction performance given the $20\times C_{max}$. In Figures \ref{fig:10x_loss_kaplan}, \ref{fig:10x_loss_chinchilla}, \ref{fig:10x_n_param_power_law}, and \ref{fig:10x_n_param_chinchilla} we illustrate a consistent prediction for $10\times$ setting compared to $20\times$ setting which is theoretically expected. Because the underlying prediction techniques model scaling behavior using power laws of the form $C_{\text{target}}^\alpha$, increasing the target compute budget by a scalar multiple effectively induces a uniform shift along the prediction trajectory in log-log space. Consequently, the relative ranking of the extrapolation techniques remains preserved, provided this shift is synchronous with the shift of the empirical optimal loss to the new target scale. To provide greater transparency for large-scale predictions, we have added new plots detailing the loss and optimal parameter ($N^*$) prediction trajectories at $10\times$ extrapolation to this section.

\subsection{Robustness and Generalizability of the Surrogate Model}

Building upon the acquisition-prediction dynamics established in Section 4.3, we conduct two targeted ablation studies to rigorously evaluate the generalizability of our framework. Our objective is to verify that the core findings, specifically the relative method rankings and the interactions between acquisition and extrapolation strategies, are intrinsic to the scaling methods themselves, rather than artifacts of the chosen surrogate model or the training data distribution.

\begin{figure*}[htbp]
    \centering
    \includegraphics[width=\textwidth]{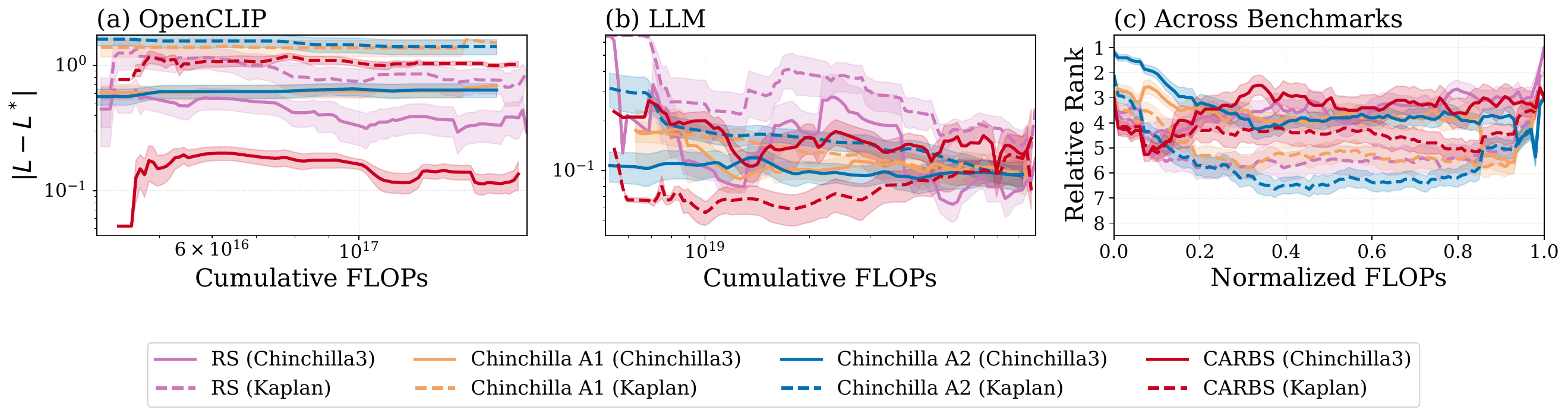}
    \caption{\textbf{Robustness of loss prediction interactions under an XGBoost surrogate.} Evaluation of acquisition and loss prediction techniques across the \textbf{(a)} OpenCLIP and \textbf{(b)} LLM benchmarks, alongside \textbf{(c)} their relative performance rankings.}
    \label{fig:20x_loss_composite_xgb}
    \vspace{1em}
    \centering
    \includegraphics[width=\textwidth]{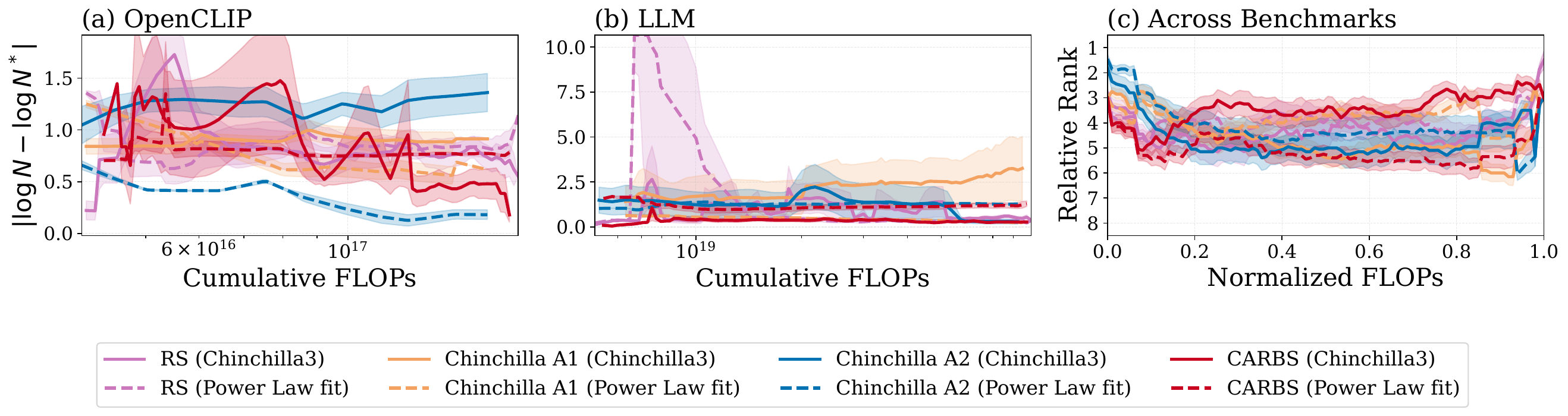}
    \caption{\textbf{Robustness of model size prediction interactions under an XGBoost surrogate.} Evaluation of acquisition and corase prediction techniques across the \textbf{(a)} OpenCLIP and \textbf{(b)} LLM benchmarks, alongside \textbf{(c)} their relative performance rankings.}
    \label{fig:20x_n_param_composite_xgb}
\end{figure*}

First, we evaluate sensitivity to the surrogate architecture. Figures \ref{fig:20x_loss_composite_xgb} and \ref{fig:20x_n_param_composite_xgb} illustrate the loss and coarse parameter extrapolation trajectories using an XGBoost surrogate in place of the primary TabPFN model. Evaluated at a highly scaled $20\times$ extrapolation budget, the XGBoost surrogate yields remarkably comparable performance trends and interaction effects, confirming that our conclusions are agnostic to the underlying surrogate mechanism.

\begin{figure*}[htbp]
    \centering
    \includegraphics[width=\textwidth]{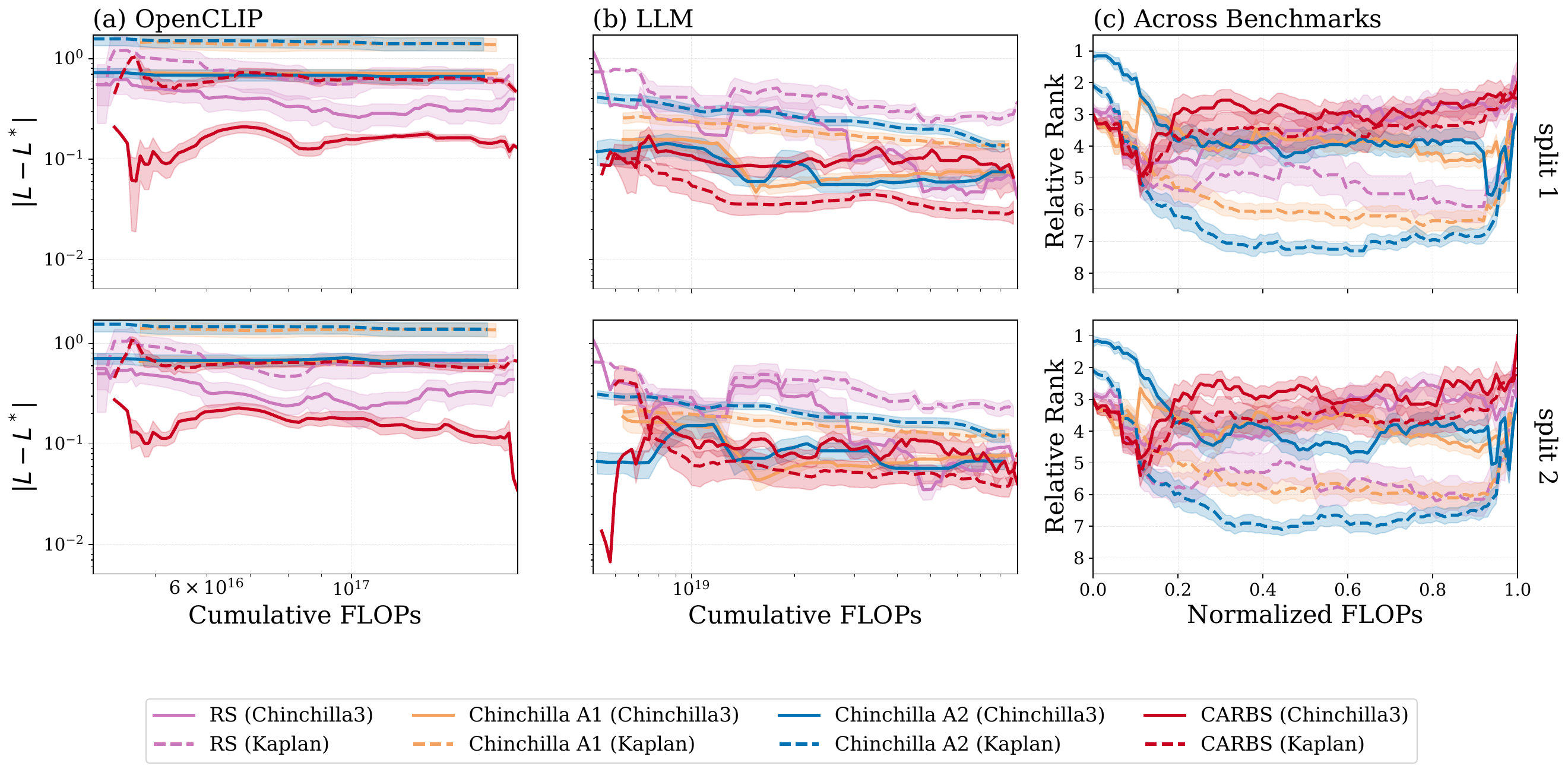}
    \caption{\textbf{Impact of disjoint data corpus splits on loss prediction interactions.} Evaluation using the TabPFN surrogate trained on independent splits illustrates that the interplay between acquisition strategies and prediction techniques (Chinchilla Approach 3 and Kaplan) remains consistent when trained on two disjoint subsets of the corpus, evaluated across the \textbf{(a)} OpenCLIP and \textbf{(b)} LLM benchmarks, alongside \textbf{(c)} their relative performance rankings.}
    \label{fig:20x_loss_composite_split}
    \vspace{1em}
    \centering
    \includegraphics[width=\textwidth]{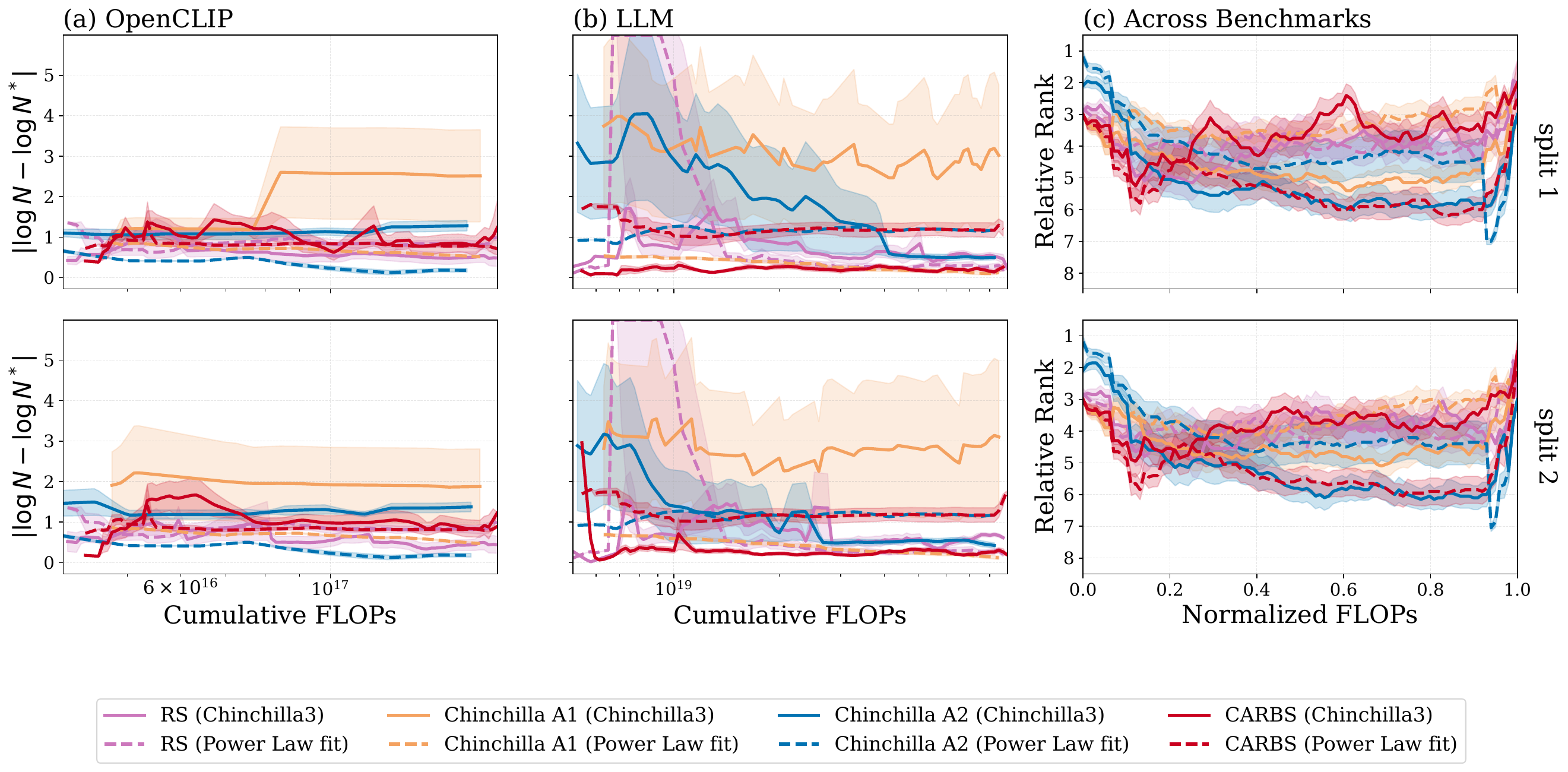}
    \caption{\textbf{Impact of disjoint data corpus splits on model size prediction interactions.} Evaluation using the TabPFN surrogate trained on independent splits illustrates that the interplay between acquisition strategies and prediction techniques (Chinchilla Approach 3 and Power-Law fit) remains consistent when trained on two disjoint subsets of the corpus, evaluated across the \textbf{(a)} OpenCLIP and \textbf{(b)} LLM benchmarks, alongside \textbf{(c)} their relative performance rankings.}
    \label{fig:20x_n_param_composite_split}
\end{figure*}

Second, we assess robustness to data variance. Figures \ref{fig:20x_loss_composite_split} and \ref{fig:20x_n_param_composite_split} present the extrapolation results when the TabPFN surrogate is trained independently on two disjoint subsets of the data corpus. Despite the shift in the empirical training distribution, the relative rankings and scaling behaviors remain strictly preserved. Together, these ablations demonstrate that the benchmarking framework's conclusions generalize reliably across both architectural substitutions and data variations.

\end{document}